\documentclass[runningheads]{llncs}

\usepackage{eccv}

\usepackage{eccvabbrv}

\usepackage{graphicx}
\usepackage{booktabs}
\usepackage{wrapfig}

\usepackage[accsupp]{axessibility}  
\usepackage{bibunits}

\usepackage[breaklinks,colorlinks,citecolor=eccvblue,hypertexnames=false]{hyperref}
\usepackage{multirow}
\usepackage[table]{xcolor}
\usepackage{float}

\usepackage{orcidlink}
\newcommand\ours{\texttt{CLIP-AU}\xspace}
\newcommand\ourstt{\texttt{CLIP-AUTT}\xspace}

\newcommand\war{\texttt{WAR}\xspace}
\newcommand\fonescore{\texttt{F1}\xspace}
\newcommand\uar{\texttt{UAR}\xspace}

\newcommand\stressid{\texttt{StressID}\xspace}

\begin{document}
\begin{bibunit}[splncs04]

\title{\ourstt: Test-Time Personalization with Action Unit Prompting for Fine-Grained Video Emotion Recognition}

\titlerunning{CLIP-AUTT}

\author{Muhammad Osama Zeeshan\inst{1}\orcidlink{0009-0006-1463-2465}  \and
Masoumeh Sharafi\inst{1} \orcidlink{0009-0001-0876-2769} \and
Benoît Savary\inst{3} \and
Alessandro Lameiras Koerich\inst{2} \orcidlink{0000-0001-5879-7014} \and
Marco Pedersoli\inst{1}\orcidlink{0000-0002-7601-8640} \and 
Eric Granger\inst{1}\orcidlink{0000-0001-6116-7945}}

\authorrunning{M.~Osama et al.}

\institute{
LIVIA, ILLS, Dept. of Systems Engineering, ETS Montreal, Canada \and
LIVIA, Dept. of Software and IT Engineering, ETS Montreal, Canada \and
Dept. of Computer Science, École Polytechnique, Paris, France \\
\email{\{muhammad-osama.zeeshan.1,masoumeh.sharafi.1\}@ens.etsmtl.ca} \\
\email{\{marco.pedersoli, alessandro.koerich, eric.granger\}@etsmtl.ca}
}
\maketitle
\begin{abstract} Personalization in emotion recognition (ER) is essential for an accurate interpretation of subtle and subject-specific expressive patterns. Recent advances in vision–language models (VLMs), such as CLIP, demonstrate strong potential for leveraging joint image–text representations in ER. However, existing CLIP-based methods either rely on CLIP’s contrastive pretraining or on LLMs to generate descriptive text prompts, which can be noisy, computationally expensive, and often fail to capture fine-grained expressions, leading to degraded performance.
In this work, Action Units (AUs) are leveraged as structured textual prompts within CLIP to model fine-grained facial expressions. AUs encode the subtle muscle activations underlying expressions, providing localized and interpretable semantic cues for more robust facial expression recognition (FER). We introduce \ours, a lightweight AU–guided temporal learning method that integrates interpretable AU semantics into CLIP. It learns generic, subject-agnostic representations by aligning AU prompts with facial dynamics, enabling fine-grained FER without CLIP fine-tuning or LLM-generated text supervision. Although \ours models fine-grained AU semantics, it does not adapt to subject-specific variability in subtle expressions. To address this limitation, we propose \ourstt, a video-based test-time personalization method that dynamically adapts AU prompts to videos from unseen subjects. By combining entropy-guided temporal window selection with prompt tuning, \ourstt enables subject-specific adaptation while preserving temporal consistency. 
Our experiments on three challenging video-based datasets — BioVid, StressID, and
BAH — indicate that \ours and \ourstt outperform state-of-the-art CLIP-based FER and TTA methods. \\
Our code: \href{https://github.com/osamazeeshan/CLIP-AUTT}{https://github.com/osamazeeshan/CLIP-AUTT}. 
\end{abstract}


\section{Introduction}
\label{sec:intro}

\begin{figure}[t!]
\centering
\includegraphics[width=1.0\linewidth]{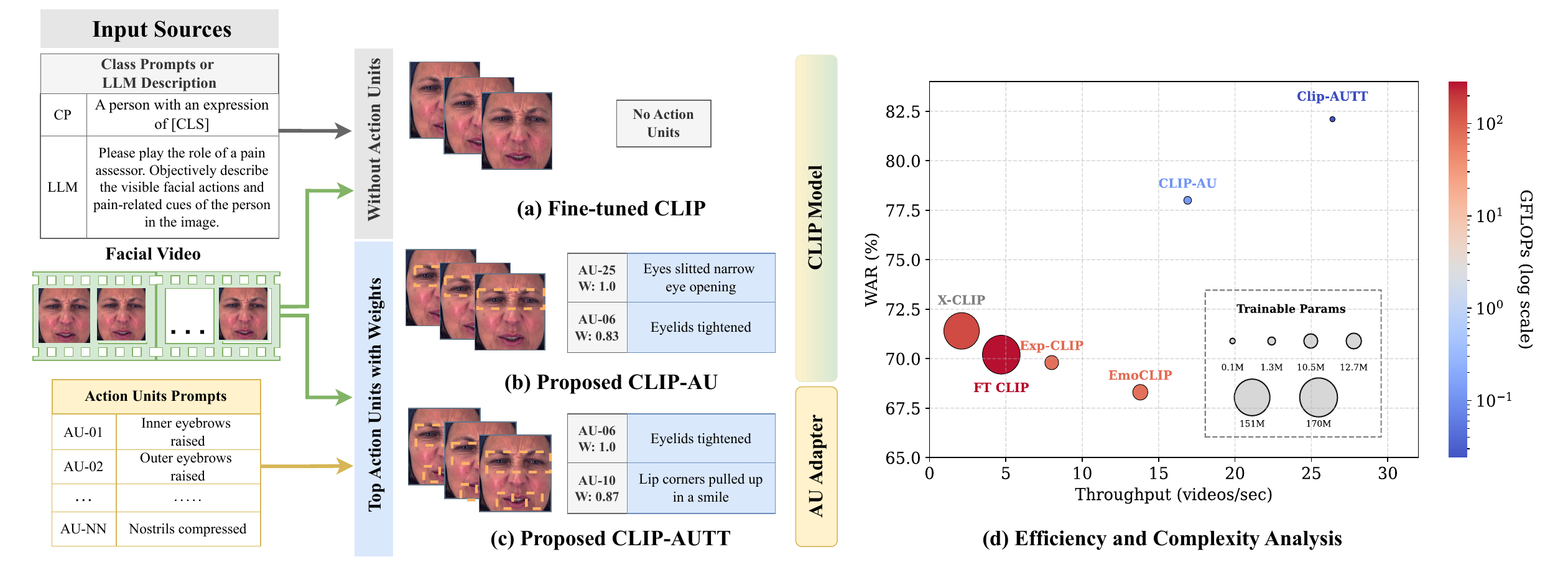}
\vspace{-5pt}
\caption{
\textbf{(a)} Existing Video ER methods based on CLIP fine-tuning use class prompts or generate LLM prompts that remain fixed at inference time, providing no subject-specific adaptation. \textbf{(b)} Our \ours introduces AU-guided pretraining to learn fine-grained expression cues. \textbf{(c)} Our \ourstt extends this for test-time personalization, combining entropy-guided temporal window selection with AU prompt tuning to adapt CLIP to each target subject. \textbf{(d)} Illustration of throughput (videos/sec), trainable parameters (M), and GFLOPs against \war. Lower GFLOPs indicate better efficiency.
}
\label{fig:clip_au_img}      
\vspace{-15pt}
\end{figure}

Large-scale vision–language models (VLMs) have demonstrated remarkable success across diverse multimodal tasks~\cite{radford2021learning, alayrac2022flamingo, li2023blip}. 
Among them, CLIP~\cite{radford2021learning} has emerged as a powerful VLM that learns a unified representation space between vision and language, enabling flexible cross-modal understanding and adaptation to downstream tasks such as image classification and retrieval~\cite{wu2023bidirectional, li2022blip, yan2023unloc}.
However, despite their versatility, CLIP-based models often struggle with specialized tasks such as recognizing subtle and fine-grained facial expressions. Unlike coarse emotion classification, subtle expressions are low-intensity and arise from localized facial muscle activations, which are difficult to capture using generic semantic prompts commonly used in CLIP-based approaches. Several recent studies have explored the use of CLIP for facial expression recognition (FER)~\cite{zhang2023learning, foteinopoulou2024emoclip, zhao2025enhancing, chen2024finecliper}, demonstrating the effectiveness of multimodal alignment for macro-level emotion classification. 
However, these approaches primarily target prototypical basic emotions and often rely on extensive fine-tuning on emotion-labeled datasets, which limits generalization and increases computational cost. 
Other methods~\cite{zhao2025enhancing, chen2024finecliper} rely on static class-prompts or large language models to generate textual prompts~(Fig.~\ref{fig:clip_au_img}a), introducing noisy or overly generic supervision that is less suitable for modeling subtle and fine-grained facial expressions.

In this paper, we leverage fine-grained Action Unit (AU) semantics within CLIP to model subtle and subject-specific facial expressions. AUs~\cite{martinez2017automatic} provide structured and interpretable descriptions of individual facial muscle activations, forming the building blocks of complex expressions~\cite{belharbi2024guided, li2023compound, liu2023facial}. 
We first introduced \ours, that incorporate AU textual descriptions directly as structured text prompts within CLIP’s text encoder.  Each AU prompt precisely describes a specific facial muscle activation (e.g., AU-06 cheek raiser, AU-25 lips part, AU-01 inner brow raiser), enabling the model to learn fine-grained visual–semantic correspondences at the muscle level as shown in Fig.~\ref{fig:clip_au_img}b.
Importantly, subtle facial expressions arise from combinations of multiple AUs, resulting in a one-to-many relationship between a single expression and several AU activations. By computing AU–video similarity scores in CLIP’s cross-modal space, \ours captures these compositional patterns without requiring AU ground-truth annotations. Instead of predicting AUs explicitly, we train a lightweight classifier on AU–video similarity features to map learned AU activation patterns directly to subtle expression categories. 
This design allows \ours to exploit structured AU semantics without AU supervision, avoiding costly CLIP fine-tuning or LLM-generated prompts while remaining computationally efficient (Fig.~\ref{fig:clip_au_img}d).

Personalization adapts models to the diverse expressive patterns of individual users~\cite{sharafi2025personalized, zeeshan2025progressive} and is essential for modeling subtle facial expressions, particularly in healthcare applications such as pain and stress assessment.
Unlike prototypical macro-emotions, subtle expressions are low-intensity, temporally evolving, and highly individualized. 
The same AU activation (e.g., AU-01 eyebrow raise) may indicate pain for one subject while reflecting a neutral expression for another. 
This subject-specific variability makes subtle expressions inherently difficult to model using generic representations.
Beyond this personalization challenge, additional variability arises from extrinsic factors such as cultural background or camera conditions~\cite{jack2009cultural}. 
While such factors affect most emotion recognition systems, subtle expressions are particularly vulnerable to misinterpretation due to their weak intensity and individualized nature. 
To address this subject-specific variability in subtle facial expressions, we introduce \textbf{\ourstt}, a video-based test-time personalization method that dynamically adapts AU prompts at inference to account for individual expressive differences (Fig.~\ref{fig:clip_au_img}c). Unlike SoTA test-time adaptation (TTA) methods for CLIP models~\cite{shu2022test,karmanov2024efficient,abdul2023align,liang2025advancing} that focus on static image classification, \ourstt explicitly models temporal dynamics to capture the gradual emergence of subtle and fine-grained facial expressions. Since peak expressive cues often appear only within specific temporal segments, we introduce an entropy-guided temporal window selection strategy to isolate the most confident and expressive window based on AU–video similarity. The selected window is then used for AU prompt tuning via entropy minimization, allowing the AU representations to adapt to how the target subject expresses the subtle emotion.
\ourstt operates on subject-specific videos, performing adaptation independently for each subject video and resetting the adapted parameters afterward to maintain stable personalization. The method performs unsupervised subject-level test-time adaptation for temporally consistent, subject-specific modeling of subtle expressions. As shown in Fig.~\ref{fig:clip_au_img}(d), under identical NVIDIA A100 48GB settings, \ourstt achieves the best performance with substantially fewer trainable parameters, lower computational cost, and higher video throughput than competing methods.

Main contributions: 
\textbf{(1)} \ours: An {AU-guided temporal learning} method is proposed that integrates AU semantics into the CLIP model, enabling it to recognize fine-grained and subtle facial expressions without requiring AU labels or LLM-generated descriptions. 
\textbf{(2)} \ourstt: A video-based test-time personalization method that dynamically adapts to unseen target subjects by optimizing AU prompts and selecting key expressive windows, enabling subject-specific modeling during inference without additional supervision.
\textbf{(3)} An extensive set of experiments was conducted to validate our methods under two settings -- (1) \ours: fine-grained video alignment and (2) \ourstt: video-based test-time personalization -- on three challenging video ER datasets -- BioVid~\cite{walter2013biovid}, StressID~\cite{chaptoukaev2023stressid}, and BAH~\cite{gonzalez2025bah} -- demonstrating that our methods outperform state-of-the-art CLIP-based FER and test-time adaptation methods.

\section{Related Work}
\label{sec:relatedwork}

\noindent\textbf{CLIP-based FER Models. }
\label{sec:LLM}
CLIP-style models learn a joint image–text embedding via contrastive pretraining on large web corpora; at inference, they score the similarity between an image and text prompts (e.g., “a photo of a [CLS] face”) to enable zero-/few-shot classification without task-specific heads~\cite{radford2021learning}. In FER, this typically means crafting emotion prompts (sometimes with prompt ensembles), optionally tuning a small set of parameters (e.g., adapters/LoRA), or enriching the text side with descriptions derived from LLMs, then classifying by the nearest text prototype. Recent works extend CLIP to FER along several axes: adapter-tuned, fine-grained prompting for dynamic FER (FineCLIPER)~\cite{chen2024finecliper}; zero-shot \emph{video} FER via sample-level captions (EmoCLIP)~\cite{foteinopoulou2024emoclip}; multimodal pretraining from verbal and nonverbal cues (EmotionCLIP)~\cite{zhang2023learning}; and LLM-augmented text prototypes with learned alignment (Exp-CLIP)~\cite{zhao2025enhancing}. Despite their zero-/few-shot positioning, these approaches still require training data.

\noindent\textbf{Facial Action Units.}
Facial Action Coding System (FACS) provides a compositional, muscle-level description of facial movements via Action Units (AUs), which many works exploit as structured priors for expression recognition \cite{ekman1978facial}. Recent AU-aware methods leverage AUs as supervision, auxiliary tasks, or intermediate reasoning cues. Li et al.\ use AU-assisted meta multi-task learning to improve compound expression recognition on RAF-CE~\cite{li2023compound}, while Liu et al.\ incorporate features from a facial AU detector into a multi-modal FER pipeline for in-the-wild videos~\cite{liu2023facial}. On the AU-detection side, FG-Net and MCM learn generalizable AU representations that support cross-corpus facial behaviour analysis~\cite{yin2024fg, zhang2024multimodal}. Beyond conventional FER, Emotion-LLaMA integrates FACS-style AU patterns into a multimodal LLM for emotion recognition and reasoning~\cite{cheng2024emotion}, and Belharbi et al.~\cite{belharbi2024guided} guide FER with a spatial AU codebook that supervises interpretable attention maps.

\noindent\textbf{Personalization in FER. }
Standard FER literature largely targets models trained and evaluated under subject-independent protocols without personalization~\cite{aslam2024distilling, li2018deep, waligora2024joint}. FER personalization is the subject-specific adaptation of a pretrained model during training, fine-tuning, or inference to mitigate subject-level distribution shifts arising from anatomical, cultural, personality, and contextual variability in real-world settings. Despite notable gains from supervised personalization methods~\cite{rescigno2020personalized, barros2019personalized}, these approaches depend on per-subject labeled data, which is costly to collect and raises privacy concerns. To address these challenges, several multi-source domain adaptation (MSDA) methods~\cite{zeeshan2024subject, zeeshan2025progressive, zeeshan2026musaco} have been proposed. These techniques leverage labeled data from multiple sources to learn domain-invariant representations for a target subject by improving robustness under cross-subject shifts. However, in many real-world applications, the source data cannot be retained or shared due to privacy constraints, which motivates source-free domain adaptation (SFDA). SFDA adapts by using only a pretrained source model and unlabeled target data via self-training, entropy minimization, and prototype refinement~\cite{sharafi2025disentangled, sharafi2025personalized}.

\noindent\textbf{Test-time Domain Adaptation.}
TTA updates a trained model at test time using unlabeled target test data captured to mitigate distribution shift, offering practical gains when source data are unavailable and full retraining is infeasible~\cite{wang2020tent, wang2024distribution, liberatori2024test, lin2023video}. While effective, conventional TTA often drifts under confirmation bias, overfits to short or non-i.i.d. streams, and depends on BatchNorm updates that fail on LayerNorm-only backbones (e.g., ViT). To address these issues, CLIP-based TTA leverages CLIP's strong zero-shot priors and text anchors to enable label-free, structure-aware adaptation through lightweight parameter updates that remain stable (via zero-init and EMA) while preserving the pretrained mapping and reducing test-time shifty~\cite{shu2022test,karmanov2024efficient,abdul2023align, liang2025advancing, zhou2022cocoop}. Loss functions aligned with CLIP contrastive pre-training allow for mitigating pseudo-label drift and class collapse~\cite{lafon2025cliptta}, prompt-ensemble strategies with periodic weight averaging yield robust updates—including single-image adaptation~\cite{osowiechi2024watt}, and backdoor analyses surface security risks for CLIP-style models~\cite{liang2024badclip}.

\section{Proposed Method}
\label{sec:methodology}
State-of-the-art CLIP-based ER approaches~\cite{zhang2023learning, foteinopoulou2024emoclip, zhao2025enhancing, chen2024finecliper} associate video frames with expression-related text prompts, typically using generic templates such as \emph{``a person with an expression of [CLS]''} or descriptions generated by large language models. While effective for macro-level emotion classification, such prompts overlook the localized facial muscle activations that characterize subtle and subject-specific expressions. Moreover, these methods often require fine-tuning of CLIP on emotion-labeled datasets, limiting generalization and increasing computational cost. Subtle facial expressions arise from specific combinations of localized facial muscle movements. To address this limitation, we propose \textbf{\ours}, which replaces class-level prompts with textual descriptions of individual AUs to explicitly model fine-grained facial muscle activations. 
Building upon this, we introduce \textbf{\ourstt}, a test-time adaptation method that enables subject-specific personalization for subtle ER.

\noindent\textbf{Preliminaries.} The proposed method builds upon a frozen VLM composed of a \textbf{visual encoder} $\mathcal{E}_v$ and a \textbf{textual encoder} $\mathcal{E}_t$, which extract visual and textual embeddings, respectively. For the textual modality, we define a set of $N = 46$ AU descriptions $\{ a_i \}_{i=1}^{N}$, where each $a_i$ is a short natural-language phrase describing a localized facial muscle movement (e.g., ``eyebrow raised'', ``nose wrinkled'', ``lip corner pulled''). The encoded AU features are further refined through a lightweight \textbf{AU adapter} $\mathcal{A}_t$ that specializes the text embeddings for expression-related semantics. The \textbf{temporal module} processes sequential frame embeddings from $\mathcal{E}_v$ and consists of a 1D convolutional encoder denoted as $\mathcal{T}(\cdot)$, a gated linear unit (GLU), and an average pooling operation to capture temporal dynamics. A two-layer \textbf{emotion classifier} $\mathcal{C}_{\text{MLP}}$ maps the AU--video similarity representations to emotion logits. Throughout this paper, $\textit{Z}_{\text{au}}$ denotes the set of AU embeddings, $\textbf{z}_v$ represents the temporally aggregated video embedding, 
$\mathbf{s}_i$ denotes the AU--window similarity vector for the $i$-th temporal segment, and $p(c|\mathbf{X})$ refers to the predicted probability for expression class $c$ for the video $\mathbf{X} \in \mathbb{R}^{T \times H \times W \times 3}$.  
All these components are consistently used across both the {\ours} and {\ourstt} methods.

\subsection{\textbf{\ours}: Fine-Grained Video Alignment}
\label{sec:au_video_alignment}
\begin{figure}[t!]
\centering
\includegraphics[width=1.0\linewidth]{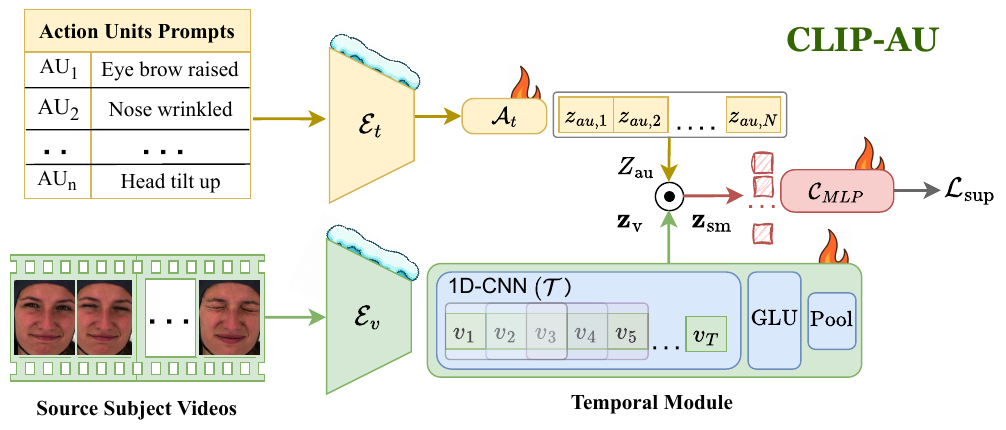}
\caption{
\textbf{Overview of the \ours.} 
The model aligns AU text embeddings with temporally encoded video representations to capture fine-grained facial dynamics for ER.
}
\label{fig:tt_au_1}      
\vspace{-10pt}
\end{figure}

During pretraining, $\mathcal{E}_t$ and $\mathcal{E}_v$ encoders are kept frozen to preserve their rich multimodal representations. A lightweight \textbf{AU adapter} ($\mathcal{A}_t$) is introduced after the $\mathcal{E}_t$ to specialize the textual embeddings toward AU-specific semantics that are critical for recognizing subtle facial expressions.  
While the frozen text encoder retains the general linguistic and visual alignment knowledge of CLIP, the AU adapter learns to extract task-relevant and fine-grained facial cues that are discriminative for emotion recognition.  
In parallel, a \textbf{temporal module} processes frame-level embeddings from the $\mathcal{E}_v$ to capture motion-aware temporal dynamics, enabling the model to encode expressive variations over time.  
Finally, an \textbf{emotion classifier} is trained to predict categorical emotions based on the learned AU features.  
The overall objective is to align these refined textual and visual representations for robust AU--video correspondence, as illustrated in Fig.~\ref{fig:tt_au_1}.

Given a source video $\mathbf{X} = \{ I_t \}_{t=1}^{T}$, each frame $I_t \in \mathbb{R}^{H \times W \times 3}$ is first encoded by the frozen image encoder $\mathcal{E}_v(\cdot)$ to obtain frame-level embeddings:
\begin{equation}
\mathbf{v}_\text{t} = \mathcal{E}_v(I_t), \quad 
{V} = [\mathbf{v}_1, \mathbf{v}_2, \ldots, \mathbf{v}_T] \in \mathbb{R}^{T \times d}.
\end{equation}
To capture the temporal information of facial expressions, sequence $V$ is processed using a 1D convolutional temporal encoder ($\mathcal{T}_{\text{}}$), followed by GLUs and an average pooling layer:
\begin{equation}
\textbf{z}_\text{v} = \text{Pool}\big(\mathrm{GLU}(\mathcal{T}(V))\big) \in \mathbb{R}^{d}.
\end{equation}
$\mathcal{T(\cdot)}$ extracts motion-aware representations by modeling local temporal continuity and fine-grained frame dependencies within a video sequence. It efficiently aggregates short-term dynamics, enabling the representation $\textbf{z}_\text{v}$ to capture subtle temporal variations in facial motion, such as transient cues linked to pain, stress, or ambivalence. $\mathrm{GLU}$ further enhances this capability by adaptively emphasizing salient temporal changes while suppressing neutral or low-variance frames, resulting in a temporally discriminative and noise-robust visual embedding.

For the textual modality, each AU description $a_i$ is encoded by the frozen text encoder $\mathcal{E}_t$ and refined through the AU adapter $\mathcal{A}_{\text{t}}(\cdot)$:
\begin{equation}
{z}_{\text{au},i} = \mathcal{A}_{\text{t}}(\mathcal{E}_t(a_i)), \quad 
\textit{Z}_{\text{au}} = [{z}_{\text{au},1}, \ldots, {z}_{\text{au},N}] \in \mathbb{R}^{N \times d}
\end{equation}
Then, cosine similarity is computed between the temporal-aware visual embedding $z_v$ and each AU embedding ${z}_{\text{au},i}$:
\begin{equation}
{z}_{\text{sm},i} = 
\frac{{z}_v^\top {z}_{\text{au},i}}
{\| {z}_v \| \, \| {z}_{\text{au},i} \|}, 
\quad 
\mathbf{z}_{\text{sm}} = [{z}_{\text{sm},1}, \ldots, {z}_{\text{sm},N}] \in \mathbb{R}^{N}
\end{equation}
The resulting similarity vector $\mathbf{z}_{\text{sm}}$ represents the activation strengths of all AUs across the video and is passed to an MLP-based emotion classifier, which is trained using a supervised loss:
\begin{equation}
\mathcal{L}_{\text{sup}} = 
- \frac{1}{M} \sum_{m=1}^{M} \sum_{c=1}^{C} y_{m,c} \log p_{m,c},
\end{equation}
where $M$ denotes the total number of training videos, $p_c$ denotes the predicted probability for class $c$ obtained from $\mathcal{C}_{\text{MLP}}(\mathbf{z}_{\text{sm}})$. This formulation enables \ours to align AU-level semantics with temporally salient visual features, facilitating robust recognition of subtle and fine-grained facial expressions.

\begin{figure*}[t!]
\centering
\includegraphics[width=1.0\linewidth]{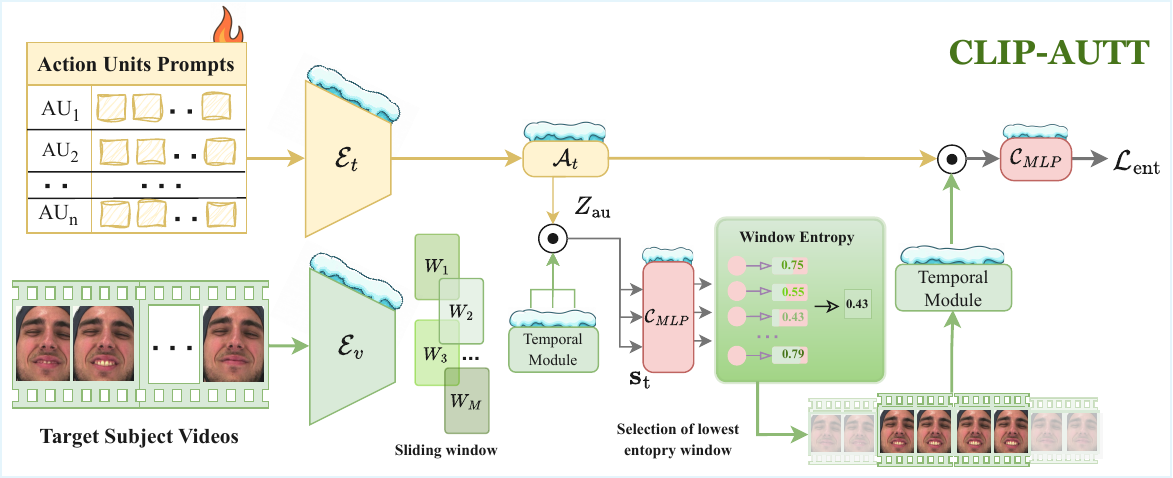}
\caption{
\textbf{Overview of the \ourstt.}  
Given a target video, \ourstt applies a sliding window with a temporal module to capture temporally coherent AU patterns and estimates window-level entropy from AU similarity to select the most expressive segment. The selected window is then used for {AU prompt tuning}, adapting AU embeddings to the target subject that results in a improve subject-specific personalized model.
}
\label{fig:tt_au_2}     
\vspace{-10pt}
\end{figure*}

\subsection{\textbf{\ourstt:} Video Test-time Personalization}
\label{sec:testtime_keyframe}
Although \ours effectively learns fine-grained AU--video alignment, its performance may degrade on unseen subjects due to differences in facial appearance, expression intensity, and motion dynamics.  
To address these challenges, we introduce \textbf{\ourstt}, a test-time adaptation method that performs unsupervised, subject-specific personalization through \textbf{Entropy-Guided Temporal Window Selection} and \textbf{AU Prompt Tuning}, as illustrated in Fig.~\ref{fig:tt_au_2}.

\subsubsection{Entropy-Guided Temporal Window Selection.}
\label{sec:testtime_keywindow}
Facial expressions associated with subtle and fine-grained emotions (e.g., pain or stress) typically evolve gradually over time, reaching a peak intensity before fading.  Consequently, not all frames in a video are equally informative; Early and late segments often contain low-intensity or ambiguous facial cues that may not reliably reflect the underlying expression.  Since each video corresponds to a single subtle expression, accurately identifying the temporally coherent peak segment is crucial for reliable recognition.  This becomes particularly important in \ourstt, as test-time AU prompt tuning relies on selecting a confident and expressive segment to accurately adapt AU semantics to the target subject. 
Adapting on weak or ambiguous segments may misguide the AU semantics and degrade personalization. 
To address this, \ourstt employs an entropy-guided temporal window selection strategy grounded in AU--video semantic alignment, focusing adaptation on the most confident and expressive window.

Given a target video
$X = \{I_t\}_{t=1}^{T}$,
each frame is first encoded using the visual encoder $\mathbf{v}_t = \mathcal{E}_v(I_t)$.
To capture the temporal evolution of subtle and fine-grained facial expressions, a sliding window of length $L$ is applied over the frame embeddings to generate
$M = T - L + 1$ overlapping windows:
\begin{equation}
W_i = \{\mathbf{v}_t\}_{t=i}^{i+L-1}, \quad i = 1, \dots, M.
\end{equation}
\noindent where $W_i$ denotes a contiguous set of $L$ frame embeddings representing a short temporal segment of the video. Each window is processed by the temporal module to capture short-term facial dynamics:
\begin{equation}
\mathbf{z}_i =
\mathrm{Pool}\big(\mathrm{GLU}(\mathcal{T}(W_i))\big)
\in \mathbb{R}^{d}.
\end{equation}
\noindent Let $\mathbf{z}_i$ denote the embedding of the $i$-th temporally aggregated window, and 
$\{\mathbf{z}_{\mathrm{au},k}\}_{k=1}^{N}$ the text embeddings of $N$ AU prompts. 
We compute the cosine similarity between $\mathbf{z}_i$ and each AU embedding and stack the resulting scores to form an AU similarity vector:
\begin{equation}
s_{i,k} =
\frac{\mathbf{z}_i^\top \mathbf{z}_{\mathrm{au},k}}
{\|\mathbf{z}_i\| \, \|\mathbf{z}_{\mathrm{au},k}\|},
\quad
\mathbf{s}_i = [s_{i,1}, \dots, s_{i,N}] \in \mathbb{R}^{N}.
\end{equation}
\noindent where $s_{i,k}$ quantifies the semantic alignment between the $i$-th window and the $k$-th AU, while $\mathbf{s}_i$ provides a structured AU-level representation of the facial expression in window $i$. The AU similarity vector is then passed to the pre-trained classifier 
$\mathcal{C}_{\mathrm{MLP}}$ to estimate expression probabilities:
\begin{equation}
\mathbf{p}_i =
\text{softmax}\big(\mathcal{C}_{\mathrm{MLP}}(\mathbf{s}_i)\big),
\end{equation}
\noindent where $\mathbf{p}_i \in \mathbb{R}^{C}$ denotes the predicted probability distribution over $C$ expression categories. To quantify prediction confidence, we compute the Shannon entropy of the window-level distribution:
\begin{equation}
h_i = - \sum_{c=1}^{C} p_i(c)\log p_i(c).
\end{equation}
\noindent A lower entropy indicates higher confidence and stronger semantic consistency between AU activations and the predicted expression. The most expressive and reliable segment is selected by $i^* = \arg\min_{i} \; h_i$.
Only the selected window $W_{i^*}$ is used for subsequent test-time AU prompt adaptation. 
By restricting personalization to the most confident window, we ensure that AU tuning is guided by temporally coherent and discriminative expression evidence.

\subsubsection{AU Prompt Tuning.} 
Recent studies have demonstrated that prompt tuning can effectively adapt large-scale VLMs to new tasks and unseen domains without full fine-tuning. Methods such as PromptAlign~\cite{abdul2023align} and TPT~\cite{shu2022test} have shown that learning or optimizing textual prompts enables efficient adaptation while preserving the pretrained model’s generalization ability.  
Motivated by these successes, we investigate whether prompt tuning can also facilitate {personalized adaptation} in {video-based facial expression recognition}, where inter-subject variability plays a crucial role. In this context, each subject expresses emotions through distinct combinations of facial muscle activations. For instance, one individual may display pain through lip stretching and eye closure, while another may exhibit it via brow raising and cheek tension. To capture these personalized patterns, \ourstt introduces {AU prompt tuning}, which optimizes Action Unit text embeddings in an unsupervised, subject-specific manner during test time to adaptively align AU semantics with individual expression styles while keeping all other modules frozen. 
  
Using the selected key window from Sec.~\ref{sec:testtime_keyframe}, their temporal embeddings are obtained through the {temporal module}, which consists of an encoder $\mathcal{T}(\cdot)$, a GLU, and an average pooling operation:
\begin{equation}
\textbf{z}_\text{v} = \text{Pool}\big(\mathrm{GLU}(\mathcal{T}(W_{i^*}))\big) \in \mathbb{R}^{d},
\end{equation}
where $W_{i^*}$ denotes the selected temporal window embeddings.  
The resulting temporal representation $\textbf{z}_\text{v}$ is then compared with the AU embeddings $\{z_{\text{au},i}\}_{i=1}^{N}$ using cosine similarity:
\begin{equation}
z_{sm,i}' = \frac{z_v^\top z_{\text{au},i}}{\| z_v \| \, \| z_{\text{au},i} \|}, 
\quad 
\mathbf{z}_\text{sm}' = [s'_1, \ldots, s'_N] \in \mathbb{R}^{N}.
\end{equation}
The similarity vector $\mathbf{z}_{sm}'$ is passed through the classifier $\mathcal{C}_{\text{MLP}}$ to produce class probabilities, and the AU prompts are optimized via entropy minimization:
\begin{equation}
\mathcal{L}_{\text{ent}} = - \sum_{c=1}^{C} p(c|\mathbf{X}) \log p(c|\mathbf{X}),
\end{equation}
where $p(c|\mathbf{X}) = \text{softmax}(\mathcal{C}_{\text{MLP}}(\mathbf{z}'_\text{sm}))_c$.
This adaptation step tunes the AU prompt embeddings $\textit{Z}_\text{au}$ to align with each distinctive subject activation pattern, improving the AU--video alignment and enhancing personalized expression recognition under unseen conditions.

\section{Results and Discussion}
\label{sec:results}

\subsection{Experimental Protocol}

\textbf{Datasets.} 
\ours was evaluated on three challenging datasets. \textbf{BioVid Heat Pain (Part A)} \cite{walter2013biovid} with 87 subjects following \cite{zeeshan2024subject}. We use 77 as the source and 10 as the target subjects. \textbf{StressID} \cite{chaptoukaev2023stressid} dataset contains recordings of different participants captured in a controlled setting while performing 11 tasks designed to trigger the stress response of each individual. We use 44 subjects as the source and 10 subjects as the target. \textbf{Behavioural Ambivalence/Hesitancy (BAH)}~\cite{gonzalez2025bah} consists of participants from diverse demographics in uncontrolled, real-world settings. We use 143 as the source and 10 subjects as the target for personalization. See suppl. materials for the complete list of target subjects.

\noindent\textbf{Implementation Details.} 
All models are optimized using the AdamW optimizer with a learning rate of $1\times10^{-3}$ and a weight decay of $1\times10^{-4}$. For \ours, we use a batch size of 8, processing eight video clips simultaneously. The AU adapter and emotion classifier are implemented as two-layer MLPs. The temporal encoder is a lightweight 1D CNN composed of a single linear layer, followed by a Gated Linear Unit (GLU) and average pooling for temporal aggregation. For \ourstt, the model operates in a video-level test-time setting, where one video is processed at a time for each subject. Thus, the batch size is set to 1. 

\begin{table*}[t]
\centering
\caption{Comparison with CLIP-based FER and TTA methods. Results are averaged over 10 target subjects per dataset. \emph{CLIP-ViT-B/32\textsuperscript{$\dagger$}} denotes full CLIP fine-tuning. \emph{ZS:} \emph{Zero-shot}; \emph{FT:} \emph{Fine-tuning}; \emph{TTA:} \emph{Test-time adaptation}.}
\vspace{0.3em}
\scriptsize
\setlength{\tabcolsep}{5pt}
\begin{tabular}{l l cccccc}
\toprule
\multirow{2}{*}{\textbf{Settings}} &
\multirow{2}{*}{\textbf{Method}} &
\multicolumn{2}{c}{\textbf{BioVid}} &
\multicolumn{2}{c}{\textbf{StressID}} &
\multicolumn{2}{c}{\textbf{BAH}} \\
\cmidrule(lr){3-4} \cmidrule(lr){5-6} \cmidrule(lr){7-8}
& & \war $\uparrow$ & \fonescore $\uparrow$ & \war $\uparrow$ & \fonescore $\uparrow$ & \war $\uparrow$ & \fonescore $\uparrow$ \\
\midrule

\multirow{1}{*}{\emph{ZS}}

& CLIP-ViT-B/32~\cite{radford2021learning}{\fontsize{4}{12} \selectfont (ICML'21)}
& 50.0 & 33.3 
& 60.4 & 34.8 
& 39.5 & 28.1 \\
\hline
\multirow{5}{*}{\emph{FT}}
& CLIP-ViT-B/32\textsuperscript{$\dagger$}~\cite{radford2021learning}{\fontsize{4}{12} \selectfont (ICML'21)}
& 69.7 & 66.6 
& \textbf{67.0} & 44.5 
& 60.4 & 39.8 \\

& EmoCLIP~\cite{foteinopoulou2024emoclip}{\fontsize{4}{12} \selectfont (FG'24)}
& 67.7 & 63.4 
& 63.5 & 35.9 
& 56.2 & 36.5 \\

& X-CLIP~\cite{ni2022expanding}{\fontsize{4}{12} \selectfont (ECCV'22)}
& 70.9 & 57.9 
& 62.3 & 41.3 
& 63.0 & 39.2 \\

& Exp-CLIP~\cite{zhao2025enhancing}{\fontsize{4}{12} \selectfont (WACV'25)}
& 70.2 & 66.7 
& 63.1 & 44.5 
& 62.2 & 38.5 \\

\rowcolor{lightgray!40}
& \textbf{\ours}
& \textbf{78.0} & \textbf{74.8} 
& 66.5 & \textbf{58.5}
& \textbf{68.3} & \textbf{40.3} \\

\midrule

\multirow{7}{*}{\emph{TTA}}

& TPT~\cite{shu2022test}{\fontsize{4}{12} \selectfont (NeurIPS'22)}
& 71.1 & 67.5 
& 70.9 & 57.9 
& 65.6 & 39.7 \\

& TDA~\cite{karmanov2024efficient}{\fontsize{4}{12} \selectfont (CVPR'24)}
& 71.4 & 68.2 
& 69.7 & 49.9 
& 65.2 & 39.9 \\

& DPE~\cite{zhang2024dual}{\fontsize{4}{12} \selectfont (NeurIPS'24)}
& 73.1 & 69.6 
& 71.3 & 54.2 
& 66.7 & 39.4 \\

& PromptAlign~\cite{abdul2023align}{\fontsize{4}{12} \selectfont (NeurIPS'23)}
& 75.3 & 71.6 
& 74.6 & 53.2 
& 67.1 & 39.7 \\

& ReTA~\cite{liang2025advancing}{\fontsize{4}{12} \selectfont (ACMMM'25)}
& 75.1 & 71.3 
& 71.8 & 52.8 
& 67.6 & 39.8 \\

& T3AL~\cite{liberatori2024test}{\fontsize{4}{12} \selectfont (CVPR'24)}
& 76.1 & 72.9 
& 75.9 & 59.4 
& 67.9 & 40.7 \\

\rowcolor{lightgray!40}
& \textbf{\ourstt}
& \textbf{81.5} & \textbf{78.0} 
& \textbf{80.8} & \textbf{77.9}
& \textbf{69.8} & \textbf{41.1} \\

\bottomrule
\end{tabular}
\label{tab:merged_results}
\vspace{-0.5em}
\end{table*}


\begin{table*}[t]
\centering
\setlength{\tabcolsep}{3pt}
\caption{\textbf{\ourstt} subject-wise comparison with TTA methods on the BioVid dataset. Best results are in bold.}
\vspace{0.3em}
\resizebox{\linewidth}{!}{
\begin{tabular}{
l
cc
cc
cc
cc
cc
cc
cc
}
\toprule
\multirow{2}{*}{\textbf{Method}} &
\multicolumn{2}{c}{\textbf{Woman-27}} &
\multicolumn{2}{c}{\textbf{Man-36}} &
\multicolumn{2}{c}{\textbf{Woman-43}} &
\multicolumn{2}{c}{\textbf{Man-25}} &
\multicolumn{2}{c}{\textbf{Woman-25}} &
\multicolumn{2}{c}{\textbf{Woman-65}} &
\multicolumn{2}{c}{\textbf{Woman-21}} \\
\cmidrule(lr){2-3}
\cmidrule(lr){4-5}
\cmidrule(lr){6-7}
\cmidrule(lr){8-9}
\cmidrule(lr){10-11}
\cmidrule(lr){12-13}
\cmidrule(lr){14-15}
& \war & \fonescore 
& \war & \fonescore 
& \war & \fonescore 
& \war & \fonescore 
& \war & \fonescore 
& \war & \fonescore 
& \war & \fonescore \\
\midrule
TPT~\cite{shu2022test}         & 91.9 & 92.0 & 53.0 & 51.5 & 61.5 & 49.6  & 88.9 & 87.0 & 76.0 & 79.5 & 98.9 & 99.0 & 79.0 & 79.0 \\
TDA~\cite{karmanov2024efficient} & 93.0 & 94.9 & 52.6 & 50.3 & 63.9 & 48.0 & 87.5 & 86.2 & 76.3 & 80.2 & 100.0 & 100.0 & 79.5 & 80.8 \\
DPE~\cite{zhang2024dual}       & 91.7 & 90.4 & 55.0 & 51.5 & 85.0 & 73.1 & 84.7 & 83.6 & 80.0 & 79.8 & 100.0 & 100.0 & 73.0 & 72.0 \\
PromptAlign~\cite{abdul2023align}  & 93.8 & 94.2 & 65.2 & 60.1 & 70.0 & 61.5 & 92.0 & 90.5 & 80.0 & 80.2 & 100.0 & 100.0 & 85.0 & 83.5 \\
ReTA~\cite{liang2025advancing} & 93.8 & 94.2 & 65.2 & 60.1 & 69.5 & 60.8 & 90.2 & 88.0 & 80.0 & 80.2 & 100.0 & 100.0 & 85.0 & 83.5 \\
T3AL~\cite{liberatori2024test} & 94.0 & \textbf{95.8} & 66.5 & 62.0 & 70.0 & 61.5 & 94.2 & 93.9 & 83.9 & 83.0 & 100.0 & 100.0 & 85.2 & 83.9 \\
\vspace{0.15em} \\[-1.5em]
\midrule
\rowcolor{gray!15}
\textbf{\ourstt} 
& \textbf{95.0} & 94.9
& \textbf{75.0} & \textbf{74.9}
& \textbf{100.0} & \textbf{100.0}
& \textbf{97.5} & \textbf{97.5}
& \textbf{92.5} & \textbf{92.4}
& \textbf{100.0} & \textbf{100.0}
& \textbf{87.5} & \textbf{87.3} \\
\bottomrule
\end{tabular}
}
\vspace{-0.5em}
\label{tab:biovid_subject_results}
\end{table*}


\begin{table*}[t]
\centering
\setlength{\tabcolsep}{3pt}
\caption{\textbf{\ourstt} subject-wise comparison with TTA methods on the StressID dataset. Best results are in bold.}
\vspace{0.3em}
\resizebox{\linewidth}{!}{%
\begin{tabular}{
l
cc
cc
cc
cc
cc
cc
cc
}
\toprule
\multirow{4}{*}{\textbf{Method}} &
\multicolumn{8}{c}{\textbf{Man}} &
\multicolumn{6}{c}{\textbf{Woman}} \\
\cmidrule(lr){2-9} 
\cmidrule(lr){10-15}
& \multicolumn{2}{c}{\textbf{Sub-1}} 
& \multicolumn{2}{c}{\textbf{Sub-2}} 
& \multicolumn{2}{c}{\textbf{Sub-3}} 
& \multicolumn{2}{c}{\textbf{Sub-4}} 
& \multicolumn{2}{c}{\textbf{Sub-5}} 
& \multicolumn{2}{c}{\textbf{Sub-6}} 
& \multicolumn{2}{c}{\textbf{Sub-7}} \\
\cmidrule(lr){2-3}
\cmidrule(lr){4-5}
\cmidrule(lr){6-7}
\cmidrule(lr){8-9}
\cmidrule(lr){10-11}
\cmidrule(lr){12-13}
\cmidrule(lr){14-15}
& \war & \fonescore 
& \war & \fonescore 
& \war & \fonescore 
& \war & \fonescore 
& \war & \fonescore 
& \war & \fonescore 
& \war & \fonescore \\
\midrule
TPT~\cite{shu2022test}          & 74.7 & 70.7 & 90.0 & 86.0 & 77.5 & 78.7 & 51.6 & 43.6  & 63.4 & 50.0 & 54.0 & 40.0 & 91.9 & 50.0 \\
TDA~\cite{karmanov2024efficient} & 66.9 & 45.5 & 75.6 & 50.1 & 80.0 & 80.0 & 45.2 & 40.0 & 75.3 & 45.6 & 81.8 & 42.8 & 91.9 & 50.0 \\
DPE~\cite{zhang2024dual}        & 64.0 & 40.9 & 73.9 & 49.0 & 80.0 & 80.0 & 55.0 & 40.6 & 73.0 & 46.0 & 81.8 & 42.8 & 92.0 & 56.6 \\
PromptAlign~\cite{abdul2023align}   & 69.9 & 41.2 & 78.6 & 45.6 & 80.0 & 80.0 & 61.6 & 45.0 & 81.3 & 58.0 & 81.8 & 42.8 & 90.9 & 45.4 \\
ReTA~\cite{liang2025advancing}  & 64.5 & 43.2 & 82.0 & 53.0 & 80.0 & 80.0 & 50.5 & 41.0 & 80.0 & 53.3 & 81.8 & 42.8 & 93.7 & 59.3 \\
T3AL~\cite{liberatori2024test}  & 71.3 & 46.3 & 80.3 & 48.6 & 80.0 & 80.0 & 62.7 & 48.6 & 77.0 & 68.0 & \textbf{82.0} & 44.0 & 93.9 & 66.0 \\
\vspace{0.15em} \\[-1.5em]
\midrule
\rowcolor{gray!15}
\textbf{\ourstt} 
& \textbf{90.9} & \textbf{90.6}
& \textbf{90.9} & \textbf{90.6}
& \textbf{100.0} & \textbf{100.0}
& \textbf{63.6} & \textbf{63.3}
& \textbf{81.8} & \textbf{80.3}
& 72.7 & \textbf{68.6}
& \textbf{100.0} & \textbf{100.0} \\
\bottomrule
\end{tabular}
}
\vspace{-0.5em}
\label{tab:stressid_subjectwise}
\end{table*}

\noindent\textbf{Baseline Methods.} 
We report results using two standard evaluation metrics. {Weighted Average Recall (\war)}, which measures the overall classification accuracy. {\fonescore-score}, which provides a balanced measure of precision and recall under imbalanced conditions.
We include two sets of baseline comparisons in our study. First, for \ours, we compare our approach against existing video-based FER methods. These include the CLIP-ViT-B/32~\cite{radford2021learning} backbone without any fine-tuning, {CLIP-ViT-B/32}\textsuperscript{$\dagger$}, which performs complete end-to-end fine-tuning of the CLIP model, and {EmoCLIP (Adapter)}~\cite{foteinopoulou2024emoclip}, where only the adapter layers are trained. We also compare with {X-CLIP}~\cite{ni2022expanding} and {Exp-CLIP}~\cite{zhao2025enhancing}, which train a projection head and leverage LLMs to extract facial expression descriptions for supervision. Second, for \ourstt, we compare against several image-based SoTA TTA methods, including {TPT}~\cite{shu2022test}, {TDA}~\cite{karmanov2024efficient}, {DPE}~\cite{zhang2024dual}, {PromptAlign}~\cite{abdul2023align}, and {ReTA}~\cite{liang2025advancing}. Further, we adapt the action recognition video-based {T3AL}~\cite{liberatori2024test} method for our classification task. To ensure a fair comparison, we use the CLIP-ViT-B/32\textsuperscript{$\dagger$} backbone, which is fine-tuned on the source subject of each dataset. 
\noindent\textbf{Prompts.} For all methods, we use the prompt \emph{``a person with an expression of [CLS]''}, which has been shown to work effectively in prior FER work~\cite{zhao2025enhancing}. 

\subsection{Comparison with State-of-the-Art Methods}
\noindent\textbf{Fine-tuning.}
Table~\ref{tab:merged_results} shows that \ours improves over all video-based CLIP FER baselines. On {BioVid} it reaches {78.0/74.8}, and on {BAH} it is best across both matrices ({68.3/40.3}). On {StressID}, EmoCLIP$^{\dagger}$ is slightly higher in \war ({67.0} vs {66.5}), but \ours achieves a decisively higher \fonescore ({58.5}, +14.0\%), indicating substantially more balanced per-class predictions under class imbalance. See suppl. materials for t-SNE analysis showing improved class separation.

\noindent\textbf{Test-time adaptation.}
Against TTA baselines in Table~\ref{tab:merged_results}, \ourstt sets a new state of the art. We compare to image-level TTA methods (TPT, TDA, DPE, PromptAlign, {ReTA}) and the video-level TTA method (T3AL). \ourstt lifts {BioVid} to {81.5/78.0}, boosts {StressID} to {80.8/77.9} (large margins in \fonescore), and improves {BAH} to {69.8/41.1}. These gains stem from temporally adapting AU prompts at test time and calibrating to each individual, which strengthens per-class balance and robustness under shift.

\noindent\textbf{Personalization.}
Tables~\ref{tab:biovid_subject_results} and \ref{tab:stressid_subjectwise} report subject-wise results on {BioVid} and {StressID}. \ourstt adapts to each person's specific AU patterns, enabling more reliable expression predictions and achieving strong performance across most subject–metric pairs. It achieves perfect scores on easier identities (e.g., BioVid \textit{Woman-43}; StressID \textit{Sub-7}) and delivers large gains on harder ones (e.g., StressID \textit{Sub-4}: \(63.6/63.3\)). Even when a baseline slightly edges \war for a subject (e.g., \textit{Sub-6}), \ourstt consistently raises \fonescore, indicating less bias toward frequent classes. Results for the BAH dataset and all \emph{10} target subjects are provided in the suppl. materials.

\subsection{Qualitative Visualization}

\begin{figure*}[t!]
\centering
\includegraphics[width=1.0\linewidth]{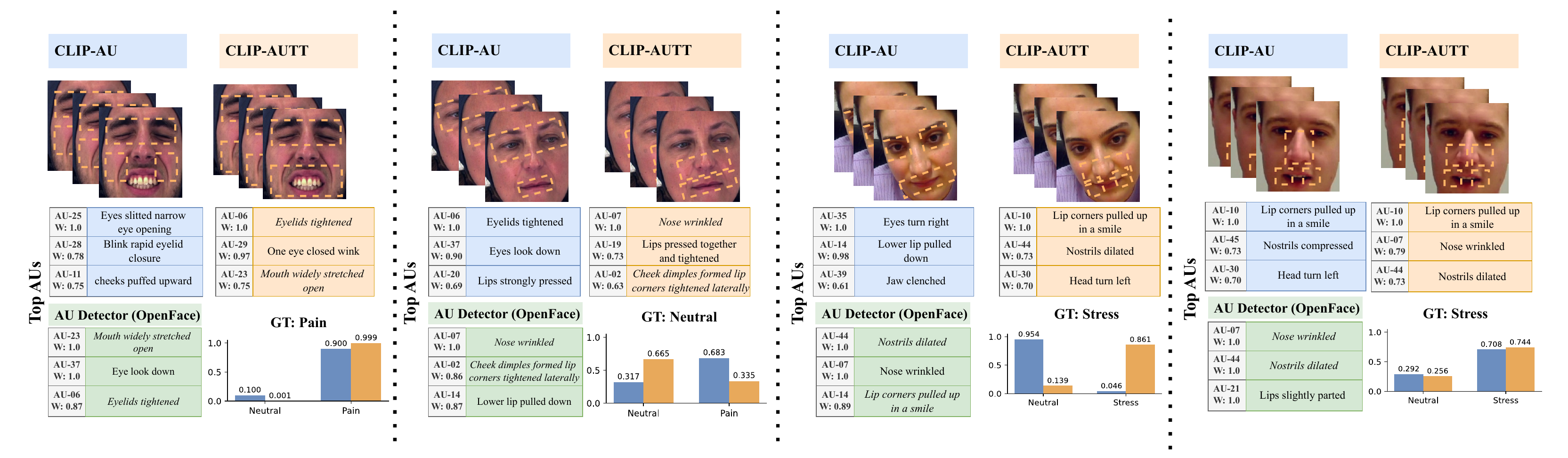}
\caption{Qualitative comparison of predicted top-activated AUs and class predictions for \ours and \ourstt, alongside AU activations estimated by an external detector (OpenFace). \ourstt shows stronger agreement with the externally estimated AUs and better alignment with the underlying expression.}
\label{fig:visualization_topau}      
\vspace{0pt}
\end{figure*}

\noindent\textbf{AU Analysis.}
Figure.~\ref{fig:visualization_topau} qualitatively compares \ours and \ourstt on subjects from the BioVid dataset. While \ours captures general facial muscle activations correlated with emotion classes, \ourstt dynamically adapts to each subject's expressive behavior, identifying more meaningful AU combinations such as fine-grained eye and mouth movements that reflect subject-specific activation patterns. Since the dataset does not provide ground-truth AU annotations, we additionally estimate AU activations using an external AU detector, Openface~\cite{baltruvsaitis2016openface}, to assess the reliability of the predicted AUs. The AU patterns predicted by \ourstt show stronger agreement with those detected by OpenFace and align better with the underlying expression, resulting in higher classification confidence. These examples demonstrate how \ours learns a generic understanding of AU semantics across subjects, while \ourstt further refines these representations through test-time adaptation to achieve personalized AU alignment without requiring AU supervision. More visualizations in suppl. material.

\noindent\textbf{Representative Temporal Window Selection.} Figure~\ref{fig:min_entropy_window} visualizes the most expressive temporal windows, their prediction uncertainty, and the minimum-entropy segment selected for adaptation. The selected window yields the most concentrated class distribution with the lowest entropy $h_3=0.493$ and, therefore, the most confident emotion prediction. Qualitatively, it also contains the clearest and most temporally consistent facial-expression cues, while the remaining windows exhibit weaker or more ambiguous evidence. Using this informative segment provides a more reliable signal for AU prompt adaptation and helps reduce updates based on uncertain frames.

\begin{figure}[t!]
\centering
\includegraphics[width=1.0\linewidth]{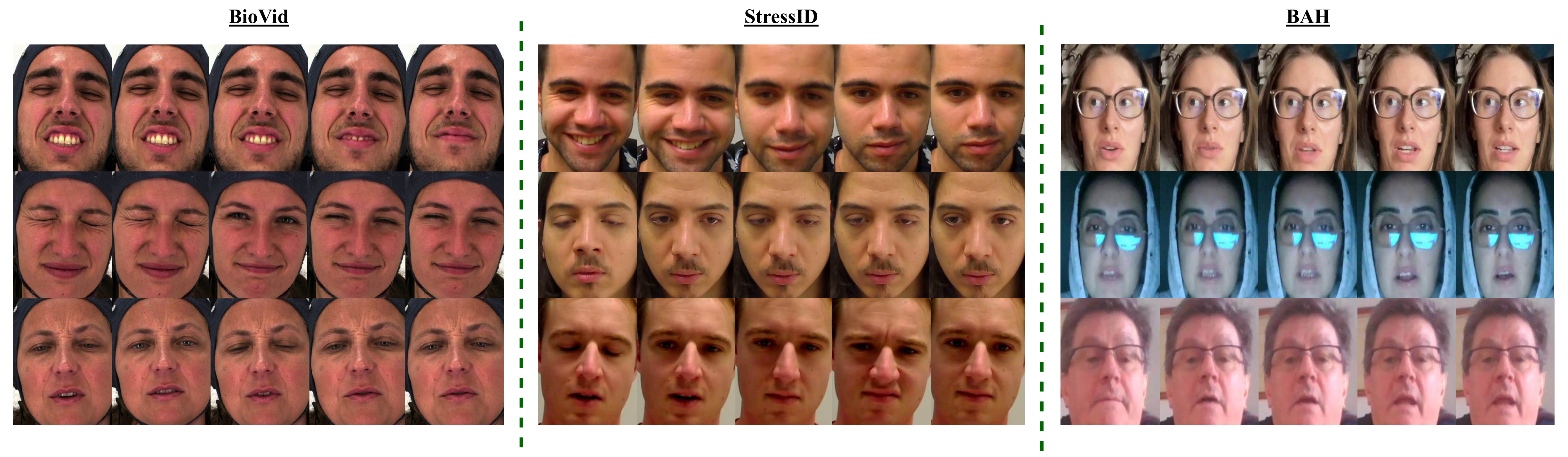}
\caption{
Samples from BioVid, StressID, and BAH highlight increased subject variability, facial movements (e.g., speaking or head motion), and environmental factors that can obscure subtle expression cues.}
\label{fig:repre_sample}    
\vspace{-10pt}
\end{figure}

\noindent\textbf{Dataset Variability Analysis.}
To better understand insights, Figure.~\ref{fig:repre_sample} visualizes representative samples from BioVid, StressID, and BAH. BioVid and StressID are collected under controlled laboratory conditions with stable lighting and minimal incidental motion. In contrast, BAH is recorded in semi-controlled settings where subjects capture videos using personal devices, resulting in larger variations in illumination, camera positioning, and incidental facial movements (e.g., speaking). These factors can obscure the low-intensity facial cues associated with subtle expressions, making reliable recognition more challenging.

\subsection{Ablation Studies}

\begin{figure*}[t]
\centering

\begin{minipage}[t]{0.44\textwidth}
\vspace{0pt}
\centering
\tiny
\renewcommand{\arraystretch}{1.1}

\begin{minipage}[c]{0.37\linewidth}
\centering
\setlength{\tabcolsep}{2.3pt}
\begin{tabular}{ccc}
\toprule
Win. & \war & \fonescore \\
\midrule
8  & 78.8 & 75.4 \\
\rowcolor{gray!25}
16 & 81.5 & 78.0 \\
24 & 80.2 & 76.8 \\
32 & 79.8 & 76.3 \\
40 & 79.0 & 75.5 \\
48 & 79.8 & 76.3 \\
56 & 80.0 & 76.6 \\
64 & 80.0 & 76.6 \\
72 & 79.0 & 75.4 \\
\bottomrule
\end{tabular}
\end{minipage}
\hfill
\begin{minipage}[c]{0.60\linewidth}
\centering
\setlength{\tabcolsep}{1.2pt}
\begin{tabular}{lcccc}
\toprule
\multirow{2}{*}{Method} &
\multicolumn{2}{c}{BioVid} &
\multicolumn{2}{c}{StressID} \\
\cmidrule(lr){2-3}
\cmidrule(lr){4-5}
& \war & \fonescore & \war & \fonescore \\
\midrule
\rowcolor{gray!10}
\multicolumn{5}{c}{\textit{Fine-tuning}} \\
Ens.\ CPs
& 69.5 & 63.9 & 61.3 & 49.9 \\
\rowcolor{gray!12}
\textbf{\ours}
& \textbf{78.0} & \textbf{74.8}
& \textbf{66.5} & \textbf{58.5} \\
\midrule
\rowcolor{gray!10}
\multicolumn{5}{c}{\textit{TTA}} \\
Ens.\ CPs
& 77.5 & 74.0 & 44.1 & 31.0 \\
\rowcolor{gray!12}
\textbf{\ourstt}
& \textbf{81.5} & \textbf{78.0}
& \textbf{80.8} & \textbf{77.9} \\
\bottomrule
\end{tabular}
\end{minipage}

\vspace{2mm}
\captionof{table}{
\textbf{Left.} Temporal-window selection: performance across window lengths ($L$). 
\textbf{Right.} Prompt-design: AU prompts versus generic class prompts under fine-tuning and TTA.
}
\label{tab:combined_analysis}
\end{minipage}
\hfill
\begin{minipage}[t]{0.54\textwidth}
\vspace{0pt}
\centering
\includegraphics[width=\linewidth]{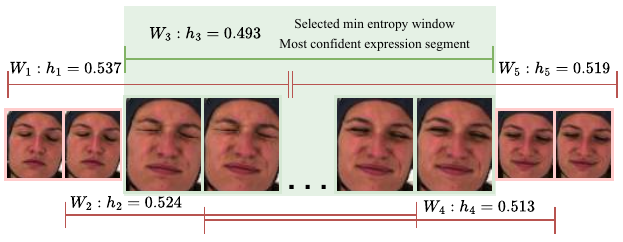}

\vspace{3.5mm}
\captionof{figure}{
Selection of the candidate temporal window with the minimum-entropy score. The selected segment contains the most confident and temporally coherent facial-expression cues that are used for personalization.}
\label{fig:min_entropy_window}
\end{minipage}
\vspace{-4mm}
\end{figure*}

\noindent\textbf{Impact of Temporal Window Selection.}
Table~\ref{tab:combined_analysis} (left) examines the impact of varying the temporal window length for \ourstt on BioVid. A shorter window length generally provides a better accuracy–efficiency trade-off, with 16 frames yielding the best performance. Importantly, \ourstt remains relatively stable across different window lengths, as the entropy-based selection reliably identifies the most expressive temporal segments. See suppl. material for extended ablation.

\noindent\textbf{Are Action Units Really Helping?}
\label{sec:res_au_helping}
To assess whether our AU adapter
truly captures subject-specific AU semantics, we replace AU prompts with an
ensemble of 46 generic FER class prompts (23 neutral and 23 emotion-specific
templates). As shown in Figure~\ref{tab:combined_analysis} (right), training with these generic CPs yields
\begin{wrapfigure}{r}{0.50\textwidth} 
  \begin{center}
    \vspace{-25pt} 
    \includegraphics[width=0.50\textwidth]{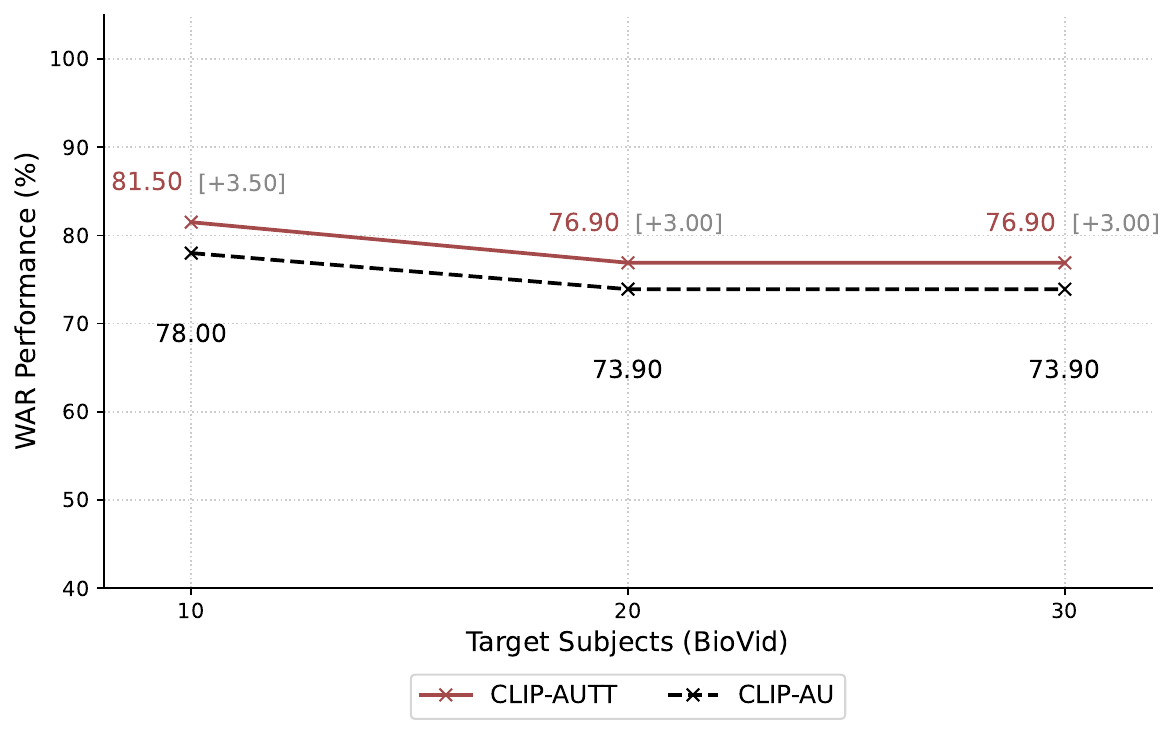}
    \vspace{-15pt}
    \caption{Analysis of robustness with a growing number of target subjects. Comparing \ours and \ourstt.}
    \label{fig:robust_tar_sub}
    \vspace{-25pt}
  \end{center}
\end{wrapfigure}
moderate performance under pre-training but collapses under TTA, most notably on \textit{StressID}, with nearly a 15\% drop. This demonstrates that class-level textual cues lack the granularity needed to model subtle, individualized expressions. In contrast, AU-based personalization consistently improves performance by enabling the adapter to align with subject-specific AU activation patterns and fine-grained facial dynamics. See the suppl. materials for CP templates and $CP+AU$ prompts ablation.

\noindent\textbf{Robustness to Subject Variability.} To evaluate robustness under increasing subject diversity, we progressively increase the number of target subjects from 10 to 30 while comparing \ourstt with the subject-agnostic baseline \ours (Figure.~\ref{fig:robust_tar_sub}). As the target pool grows, performance naturally reflects the increased subject heterogeneity and the reduced number of source subjects available for training. Importantly, the performance trend remains smooth and consistent, indicating stable generalization. Across all target sizes, \ourstt consistently outperforms \ours by about three \war points, demonstrating reliable personalization benefits independent of the target pool size.

\begin{wraptable}{r}{0.5\linewidth}
\vspace{-1.2em}
\centering
\scriptsize
\setlength{\tabcolsep}{4.5pt}
\begin{tabular}{cccc}
\toprule
Win. Sel. & AU Tune & \war & \fonescore \\
\midrule
            &            & 78.0 & 74.8 \\
\checkmark  &            & 78.7 & 75.2 \\
            & \checkmark & 80.0 & 76.4 \\
\checkmark  & \checkmark & \textbf{81.5} & \textbf{78.0} \\
\bottomrule
\end{tabular}
\vspace{-0.2em}
\caption{Analysis of individual components on the BioVid dataset.}
\label{tab:component_ablation}
\vspace{-2.0em}
\end{wraptable}
\noindent\textbf{Individual Component Analysis.}
As shown in Table~\ref{tab:component_ablation}. We isolate the contributions of minimum-entropy window selection and AU prompt tuning. Starting from the \ours baseline, window selection alone provides modest gains of $0.7\%$ \war and $0.4\%$ \fonescore, while AU prompt tuning yields larger improvements of $2.0\%$ \war and $1.6\%$ \fonescore. Combining both components achieves the best performance, improving the baseline by $3.5\%$ and $3.2\%$ for \war and \fonescore, respectively. These results show that all the components were significantly informative, while the AU prompt tuning is the main contributor to guiding the adaptation process.

\section{Conclusion}
\label{sec:conclusion}
Our study highlights a key challenge in facial expression recognition by capturing the individuality and variability of human emotional expression. While existing CLIP-based ER models rely on generic text prompts or LLM-generated descriptions to align visual features with emotion categories, they fail to model the localized AU-level cues and the diversity of subject-specific facial behaviors. Motivated by this limitation, we introduced two approaches: \ours, which integrates fine-grained AU-aware representations, and \ourstt, which adapts AU prompts to each subject at test time. Our approach learns distinctive AU activation patterns for each subject, enabling adaptive modeling of individual expressive traits. For instance, pain may manifest through “eye tightening’’ and “mouth opening’’ in some individuals, whereas others exhibit cues such as “cheek raising’’ or “rapid eyelid closure.’’ By adapting AU semantics to each subject, our method learns these personalized activation patterns without requiring subject-level supervision, enabling robust and fine-grained emotion recognition. 

\noindent\textbf{Supplementary materials.}
Includes subject-wise TTA results, baseline details, extended ablations on temporal design, AU adapters, CP–AU fusion, and window selection, plus qualitative analyses.

\section*{Acknowledgment} This research was partially supported by the Natural Sciences and Engineering Research Council of Canada, Fonds de recherche du Québec – Santé, Canada Foundation for Innovation, and the Digital Research Alliance of Canada.

%
%
\putbib[main]
\end{bibunit}

\clearpage
\setcounter{page}{1}
\begin{bibunit}[splncs04]

\setcounter{section}{0}
\setcounter{subsection}{0}
\setcounter{subsubsection}{0}
\setcounter{figure}{0}
\setcounter{table}{0}
\setcounter{equation}{0}

\renewcommand{\thesection}{\arabic{section}}
\renewcommand{\thesubsection}{\arabic{section}.\arabic{subsection}}
\renewcommand{\thesubsubsection}{\arabic{section}.\arabic{subsection}.\arabic{subsubsection}}
\renewcommand{\thefigure}{\arabic{figure}}
\renewcommand{\thetable}{\arabic{table}}
\renewcommand{\theequation}{\arabic{equation}}

\makeatletter
\renewcommand{\theHsection}{supp.\arabic{section}}
\renewcommand{\theHsubsection}{supp.\arabic{section}.\arabic{subsection}}
\renewcommand{\theHsubsubsection}{supp.\arabic{section}.\arabic{subsection}.\arabic{subsubsection}}
\renewcommand{\theHfigure}{supp.\arabic{figure}}
\renewcommand{\theHtable}{supp.\arabic{table}}
\renewcommand{\theHequation}{supp.\arabic{equation}}
\makeatother

\begin{center}
{\Large\bfseries
\ourstt: Test-Time Personalization with Action Unit Prompting for\\
Fine-Grained Video Emotion Recognition\par}
\vspace{0.35em}
{\large\bfseries Supplementary Materials\par}
\vspace{0.8em}

{\normalsize
Muhammad Osama Zeeshan\textsuperscript{1}\orcidlink{0009-0006-1463-2465},
Masoumeh Sharafi\textsuperscript{1}\orcidlink{0009-0001-0876-2769},
Beno\^it Savary\textsuperscript{3},\\
Alessandro Lameiras Koerich\textsuperscript{2}\orcidlink{0000-0001-5879-7014},
Marco Pedersoli\textsuperscript{1}\orcidlink{0000-0002-7601-8640}, and
Eric Granger\textsuperscript{1}\orcidlink{0000-0001-6116-7945}\par}
\vspace{0.55em}

{\small
\textsuperscript{1}LIVIA, ILLS, Dept. of Systems Engineering, ETS Montreal, Canada\\
\textsuperscript{2}LIVIA, Dept. of Software and IT Engineering, ETS Montreal, Canada\\
\textsuperscript{3}Dept. of Computer Science, \textit{\'Ecole Polytechnique}, Paris, France\par}
\vspace{0.45em}

{\small
\texttt{\{muhammad-osama.zeeshan.1,masoumeh.sharafi.1\}@ens.etsmtl.ca}\\
\texttt{\{marco.pedersoli,alessandro.koerich,eric.granger\}@etsmtl.ca}\\
\texttt{\href{https://osamazeeshan.github.io/publications/zeeshan2026clipautt}{https://osamazeeshan.github.io/CLIP-AUTT}}\par}
\end{center}

\vspace{0.8em}
\noindent\textbf{Contents}\par
\vspace{0.3em}
{\small
\setlength{\parskip}{0pt}
\newcommand{\suppcontentsline}[3]{%
  \noindent\hspace*{#1}%
  \hyperref[#2]{\ref*{#2}\enspace #3}%
  \dotfill\pageref*{#2}\par}
\suppcontentsline{0em}{sec:supp_prompt_tuning}{Explanation of AU Prompt Tuning for Personalization}
\suppcontentsline{0em}{sec:supp_experiments}{Additional Experimental Details}
\suppcontentsline{1.5em}{subsec:supp_impl}{Implementation Details and TTA Baseline Methods}
\suppcontentsline{1.5em}{subsec:supp_targets}{Target Subjects for \ours and \ourstt}
\suppcontentsline{1.5em}{subsec:supp_aus}{Action Units Used in \ourstt}
\suppcontentsline{1.5em}{subsec:supp_personalization}{Additional Personalization Results}
\suppcontentsline{0em}{sec:supp_ablation}{Ablation}
\suppcontentsline{1.5em}{subsec:supp_temporal}{Temporal Module Design Choice}
\suppcontentsline{1.5em}{subsec:supp_class_template}{Class-Prompt Template Selection}
\suppcontentsline{1.5em}{subsec:supp_class_replace}{Replacing AU Prompts with Class Prompts}
\suppcontentsline{1.5em}{subsec:supp_cp_au}{Effect of Adding Class Prompts with AU Prompts}
\suppcontentsline{1.5em}{subsec:supp_adapter}{Impact of AU Adapter}
\suppcontentsline{1.5em}{subsec:supp_keyframe}{Extended Key-Frame Selection Analysis}
\suppcontentsline{1.5em}{subsec:supp_additional}{Additional Analysis of \ourstt}
}
\vspace{1em}

\section{Explanation of AU Prompt Tuning for Personalization}
\label{sec:supp_prompt_tuning}

To enable subject-specific personalization without requiring AU annotations, \ourstt\ optimizes the AU text embeddings using an entropy-minimization objective. Given a target video, the model first computes AU--video cosine similarities, produces an emotion probability distribution through a lightweight classifier, and then minimizes the prediction entropy:
\[
\mathcal{L}_{\text{ent}} = -\sum_{c} p(c|X)\log p(c|X),
\]
which encourages the model to make confident and consistent predictions for that subject. Importantly, only the AU embeddings $z_{au,i}$ are updated during adaptation; all CLIP encoders and classifier weights remain frozen.

Formally, minimizing $\mathcal{L}_{\text{ent}}$ updates the AU embeddings $z_{au,i}$ while keeping the classifier weights $\mathbf{M}$ fixed. Each row of $\mathbf{M}$ encodes the contribution of individual AUs to each emotion class, and the corresponding entropy gradient
\[
\nabla_{z_{au,i}}\mathcal{L}_{\text{ent}} \propto 
\Big(\sum_c p_c \mathbf{M}_{c,i} - \mathbf{M}_{y^*,i}\Big)
\frac{\partial \cos(z_v, z_{au,i})}{\partial z_{au,i}},
\]
indicates that AUs with high positive weights for the most probable class ($\mathbf{M}_{y^*,i}$ large) are pulled toward the video embedding $z_v$, increasing their cosine similarity. Conversely, AUs with low relevance or contradictory influence are pushed away. This process sharpens the predicted class distribution by amplifying class-relevant AUs and suppressing uninformative ones, effectively re-weighting AU importance for the target subject video. Geometrically, entropy minimization rotates and adjusts AU embeddings in the joint CLIP embedding space so that class-diagnostic AUs (e.g., eye tightening or mouth stretch for pain) align more closely with the subject's visual embedding. This enables personalized AU video alignment and improved recognition performance, all without requiring AU supervision.

\section{Additional Experimental Details}
\label{sec:supp_experiments}

\subsection{Implementation Details and TTA Baseline Methods}
\label{subsec:supp_impl}

All experiments were conducted on an NVIDIA A100 GPU (48 GB). 
For \ours, the pre-training stage was performed for 10 epochs. 
For \ourstt, test-time adaptation is applied at the subject level, where each video undergoes 10 iterations.
We compare with six CLIP–based test-time adaptation baselines. \textbf{TPT}~\cite{shu2022test} optimizes only textual prompt tokens at inference via entropy minimization over strongly augmented views, keeping both encoders frozen. 
\textbf{TDA}~\cite{karmanov2024efficient} is a training-free dynamic adapter that maintains a feature–label cache, performs progressive pseudo-label refinement (including negative pseudo-labels), and fuses cache logits with CLIP zero-shot logits. \textbf{DPE}~\cite{zhang2024dual} jointly evolves textual and visual class prototypes online, adds light per-sample residuals, and enforces cross-modal alignment so the two prototype spaces remain consistent under shift. \textbf{PromptAlign}~\cite{abdul2023align} adapts prompts by explicitly aligning test-time feature statistics to source statistics to reduce distribution mismatch beyond plain entropy objectives. \textbf{ReTA}~\cite{liang2025advancing} targets reliability in cache-based VLM TTA through CER (consistency-aware entropy reweighting) for selective cache updates and DDC (diversity-driven distribution calibration) that models each class as a Gaussian family over evolving text embeddings to yield calibrated decision boundaries. Finally, \textbf{T3AL}~\cite{liberatori2024test}, originally for zero-shot temporal action localization in videos, is adapted for classification by combining video-level pseudo-labeling from aggregated frame features, self-supervised refinement of frame-wise scores, and text-guided region suppression, with encoders frozen and updates reset per sample.

\subsection{Target Subjects for \textbf{\ours} and \textbf{\ourstt}}
\label{subsec:supp_targets}
We evaluate test-time personalization on three datasets. \textbf{BioVid}~\cite{walter2013biovid},
following prior work~\cite{zeeshan2025progressive, zeeshan2026musaco}, we evaluate on 10 target subjects:
\texttt{081014\_w\_27},  
\texttt{101609\_m\_36},  
\texttt{112009\_w\_43},  
\texttt{091809\_w\_43},  
\texttt{071309\_w\_21},  
\texttt{073114\_m\_25},  
\texttt{080314\_w\_25},

\noindent\texttt{073109\_w\_28},  
\texttt{100909\_w\_65},  
\texttt{081609\_w\_40}.  
These subjects represent a diverse set of individual pain-expression patterns relevant for personalization. \textbf{StressID}~\cite{chaptoukaev2023stressid}, we follow the protocol in prior work ~\cite{zeeshan2026musaco} to select 10 target subjects:
\texttt{kycf},  
\texttt{uymz},  
\texttt{h8s1},  
\texttt{ctzy},  
\texttt{p9i3},  
\texttt{7h5u},  
\texttt{g7r2},  
\texttt{b9w0},  
\texttt{r3zm},  
\texttt{x1q3}.  
These identities show substantial variability in subtle stress-related facial behaviors, making them suitable for evaluating personalization. \textbf{BAH}~\cite{gonzalez2025bah}, we also select 10 target subjects:
\texttt{82711},  
\texttt{82687},  
\texttt{82585},  
\texttt{82592},  
\texttt{82598},  
\texttt{82632},  
\texttt{82681},  
\texttt{82683},  
\texttt{82708},  
\texttt{82714}.  
These subjects span a broad range of demographic and expression variations, enabling a comprehensive evaluation of subtle video-based ER.

\begin{table*}[t]
\centering
\setlength{\tabcolsep}{3.5pt}
\renewcommand{\arraystretch}{1.1}
\caption{Complete list of 46 Action Unit (AU) textual prompts used in \ours. Each AU corresponds to a distinct facial muscle movement or expression cue.}
\resizebox{\linewidth}{!}{%
\begin{tabular}{llll}
\toprule
\textbf{AU ID} & \textbf{Description} & \textbf{AU ID} & \textbf{Description} \\
\midrule
AU01 & inner\_eyebrows\_raised & AU24 & upper\_eyelids\_drooping \\
AU02 & outer\_eyebrows\_raised & AU25 & eyes\_slitted\_narrow\_eye\_opening \\
AU03 & eyebrows\_lowered\_and\_pulled\_together & AU26 & eyes\_fully\_closed \\
AU04 & upper\_eyelids\_lifted & AU27 & squinting\_lower\_and\_upper\_lids\_compressed \\
AU05 & cheeks\_raised\_and\_eye\_corners\_tightened & AU28 & blink\_rapid\_eyelid\_closure \\
AU06 & eyelids\_tightened & AU29 & one\_eye\_closed\_wink \\
AU07 & nose\_wrinkled & AU30 & head\_turn\_left \\
AU08 & upper\_lip\_lifted & AU31 & head\_turn\_right \\
AU09 & nasolabial\_folds\_deepened & AU32 & head\_tilt\_up \\
AU10 & lip\_corners\_pulled\_up\_in\_a\_smile & AU33 & head\_tilt\_down \\
AU11 & cheeks\_puffed\_upward & AU34 & eyes\_turn\_left \\
AU12 & cheek\_dimples\_formed\_lip\_corners\_tightened\_laterally & AU35 & eyes\_turn\_right \\
AU13 & lip\_corners\_pulled\_down & AU36 & eyes\_look\_up \\
AU14 & lower\_lip\_pulled\_down & AU37 & eyes\_look\_down \\
AU15 & chin\_pushed\_up\_lower\_lip\_bulged & AU38 & jaw\_thrust\_forward \\
AU16 & lips\_puckered\_forward & AU39 & jaw\_clenched \\
AU17 & lips\_stretched\_horizontally & AU40 & lip\_bite \\
AU18 & lips\_funneled\_rounded\_and\_protruded & AU41 & lip\_wipe\_with\_tongue \\
AU19 & lips\_pressed\_together\_and\_tightened & AU42 & lick\_lips \\
AU20 & lips\_strongly\_pressed & AU43 & cheek\_blow \\
AU21 & lips\_slightly\_parted & AU44 & nostrils\_dilated \\
AU22 & jaw\_lowered\_mouth\_opened & AU45 & nostrils\_compressed \\
AU23 & mouth\_widely\_stretched\_open & AU46 & tongue\_show \\
\bottomrule
\end{tabular}
}
\label{tab:au_prompts_full}
\end{table*}

\begin{table*}[t]
\centering
\setlength{\tabcolsep}{3pt}
\renewcommand{\arraystretch}{1.15}

\caption{Subject-wise performance across 10 BioVid target subjects. Metrics are grouped into \war, \uar, and \fonescore.}

\resizebox{\linewidth}{!}{%
\begin{tabular}{lccccccccccc}
\toprule
\textbf{Method} & Sub-1 & Sub-2 & Sub-3 & Sub-4 & Sub-5 & Sub-6 & Sub-7 & Sub-8 & Sub-9 & Sub-10 & Avg \\
\midrule
\multicolumn{12}{l}{\textbf{\war (\%)}} \\
\midrule
TPT          & 91.9 & 53.0 & 50.0 & 61.5 & 79.0 & 88.9 & 76.0 & 50.0 & 98.9 & 62.0 & 71.1 \\
TDA          & 93.0 & 52.6 & 50.0 & 63.9 & 79.5 & 87.5 & 76.3 & 50.0 & 100.0 & 61.6 & 71.4 \\
DPE          & 91.7 & 55.0 & 50.0 & 85.0 & 73.0 & 84.7 & 80.0 & 50.0 & 100.0 & 61.8 & 73.1 \\
PromptAlign  & 93.8 & 65.2 & 50.0 & 70.0 & 85.0 & 92.0 & 80.0 & 50.0 & 100.0 & 67.6 & 75.3 \\
ReTA         & 93.8 & 65.2 & 50.0 & 69.5 & 85.0 & 90.2 & 80.0 & 50.0 & 100.0 & 67.6 & 75.1 \\
T3AL         & 94.0 & 66.5 & 50.0 & 70.0 & 85.2 & 94.2  & 83.9 & 50.0 & 100.0 & 67.6 & 76.1 \\
\rowcolor{gray!15}
\textbf{\ourstt} & {95.0} & {75.0} & 50.0 & {100.0} & 92.5 & {97.5} & {92.5} & 50.0 & 100.0 & 62.5 & \textbf{81.5} \\
\midrule
\multicolumn{12}{l}{\textbf{\uar (\%)}} \\
\midrule
TPT          & 91.9 & 53.0 & 50.0 & 61.5 & 79.0 & 88.9 & 76.0 & 50.0 & 98.9 & 62.0 & 71.1 \\
TDA          & 93.0 & 52.6 & 50.0 & 63.9 & 79.5 & 87.5 & 76.3 & 50.0 & 100.0 & 61.6 & 71.4 \\
DPE          & 91.7 & 55.0 & 50.0 & 85.0 & 73.0 & 84.7 & 80.0 & 50.0 & 100.0 & 61.8 & 73.1 \\
PromptAlign  & 93.8 & 65.2 & 50.0 & 70.0 & 85.0 & 92.0 & 80.0 & 50.0 & 100.0 & 67.6 & 75.4 \\
ReTA         & 93.8 & 65.2 & 50.0 & 69.5 & 85.0 & 90.2 & 80.0 & 50.0 & 100.0 & 67.6 & 75.1 \\
T3AL         & 94.0 & 66.5 & 50.0 & 70.0 & 85.2 & 94.2 & 83.9 & 50.0 & 100.0 & 67.6 & 76.1 \\
\rowcolor{gray!15}
\textbf{\ourstt} & {95.0} & {75.0} & 50.0 & {100.0} & 92.5 & {97.5} & {92.5} & 50.0 & 100.0 & 62.5 & \textbf{81.5} \\
\midrule
\multicolumn{12}{l}{\textbf{\fonescore (\%)}} \\
\midrule
TPT          & 92.0 & 51.5 & 40.0 & 49.6 & 79.0 & 87.0 & 79.5 & 43.4 & 99.0 & 54.0 & 67.5 \\
TDA          & 94.9 & 50.3 & 43.5 & 48.0 & 80.8 & 86.2 & 80.2 & 43.5 & 100.0 & 55.0 & 68.2 \\
DPE          & 90.4 & 51.5 & 43.5 & 73.1 & 72.0 & 83.6 & 79.8 & 48.5 & 100.0 & 54.4 & 69.6 \\
PromptAlign  & 94.2 & 60.1 & 42.5 & 61.5 & 83.5 & 90.5 & 80.2 & 43.5 & 100.0 & 60.0 & 71.6 \\
ReTA         & 94.2 & 60.1 & 43.5 & 60.8 & 83.5 & 88.0 & 80.2 & 43.5 & 100.0 & 60.0 & 71.3 \\
T3AL         & 95.8 & 62.0 & 43.9 & 61.5 & 83.9 & 93.9 & 83.0 & 43.5 & 100.0 & 62.0 & 72.9 \\
\rowcolor{gray!15}
\textbf{\ourstt} & 95.0 & {73.9} & 33.3 & {100.0} & 92.5 & {97.5} & {92.4} & 33.3 & 100.0 & 62.4 & \textbf{78.0} \\
\bottomrule
\end{tabular}
}
\label{tab:biovid_subject_full_metrics}
\end{table*}

\begin{table*}[t]
\centering
\setlength{\tabcolsep}{3pt}
\renewcommand{\arraystretch}{1.15}

\caption{{Subject-wise performance across 10 StressID target subjects.}
Metrics are grouped into \war, \uar, and \fonescore}
\vspace{0.3em}
\resizebox{\linewidth}{!}{%
\begin{tabular}{lccccccccccc}
\toprule
\textbf{Method} & Sub-1 & Sub-2 & Sub-3 & Sub-4 & Sub-5 & Sub-6 & Sub-7 & Sub-8 & Sub-9 & Sub-10 & Avg \\
\midrule
\multicolumn{12}{l}{\textbf{\war (\%)}} \\
\midrule
TPT & 74.7 & 51.6 & 90.0 & 88.0 & 52.2 & 77.5 & 66.6 & 63.4 & 54.0 & 91.9 & 70.9 \\ TDA & 66.9 & 45.2 & 75.6 & 90.1 & 49.6 & 80.0 & 41.2 & 75.3 & 81.8 & 91.9 & 69.7 \\ DPE & 64.0 & 55.0 & 73.9 & 90.0 & 52.6 & 80.0 & 50.9 & 73.0 & 81.8 & 92.0 & 71.3 \\ PromptAlign & 69.9 & 61.6 & 78.6 & 90.9 & 51.2 & 80.0 & 60.4 & 81.3 & 81.8 & 90.9 & 74.6 \\ ReTA & 64.5 & 50.5 & 82.0 & 90.0 & 46.8 & 80.0 & 49.3 & 80.0 & 81.8 & 93.7 & 71.8 \\ T3AL & 71.3 & 62.7 & 80.3 & 90.2 & 54.0 & 80.0 & 68.0 & 77.0 & 82.0 & 93.9 & 75.9 \\
\rowcolor{gray!15}
\textbf{\ourstt} & {90.9} & {63.6} & {90.9} & 81.0 & 54.6 & 100.0 & {72.7} & 81.8 & 72.7 & {100.0} & \textbf{80.8} \\
\midrule
\multicolumn{12}{l}{\textbf{\uar (\%)}} \\
\midrule
TPT          & 70.0 & 55.0 & 90.1 & 79.6 & 59.0 & 77.5 & 50.0 & 54.0 & 52.6 & 56.0 & 64.4 \\
TDA          & 53.0 & 52.0 & 55.9 & 76.8 & 58.4 & 80.0 & 54.5 & 50.0 & 51.0 & 56.0 & 58.8 \\
DPE          & 55.4 & 54.8 & 54.9 & 76.9 & 57.0 & 80.0 & 51.0 & 51.0 & 50.0 & 67.0 & 59.8 \\
PromptAlign  & 55.0 & 54.0 & 61.0 & 81.0 & 53.0 & 80.0 & 60.2 & 65.7 & 50.0 & 50.0 & 61.0 \\
ReTA         & 54.3 & 51.4 & 61.2 & 75.0 & 51.0 & 80.0 & 55.0 & 61.9 & 50.0 & 61.0 & 60.1 \\
T3AL         & 56.7 & 60.0 & 62.8 & 82.9 & 54.4 & 80.0 & 62.0 & 77.6 & 54.9 & 65.0 & 65.6 \\
\rowcolor{gray!15}
\textbf{\ourstt} & {90.0} & {75.0} & {92.9} & 88.9 & 58.3 & 100.0 & 60.4 & 80.3 & 70.8 & {100.0} & \textbf{81.7} \\
\midrule
\multicolumn{12}{l}{\textbf{\fonescore (\%)}} \\
\midrule
TPT          & 70.7 & 43.6 & 86.0 & 83.1 & 36.4 & 78.7 & 41.1 & 50.0 & 40.0 & 50.0 & 57.9 \\
TDA          & 45.5 & 40.0 & 50.1 & 81.0 & 35.0 & 80.0 & 29.6 & 45.6 & 42.9 & 50.0 & 49.9 \\
DPE          & 41.0 & 40.6 & 49.0 & 81.0 & 49.8 & 80.0 & 56.0 & 46.0 & 42.9 & 56.6 & 54.2 \\
PromptAlign  & 41.3 & 45.0 & 45.6 & 82.3 & 48.1 & 80.0 & 44.0 & 58.0 & 42.9 & 45.5 & 53.2 \\
ReTA         & 43.2 & 41.0 & 53.0 & 80.9 & 36.1 & 80.0 & 39.1 & 53.3 & 42.9 & 59.3 & 52.8 \\
T3AL         & 46.3 & 48.6 & 48.6 & 80.1 & 51.0 & 80.0 & 61.8 & 68.0 & 44.0 & 66.0 & 59.4 \\
\rowcolor{gray!15}
\textbf{\ourstt} & {90.6} & {63.3} & {90.6} & 77.0 & 47.6 & 100.0 & 61.1 & 80.3 & 68.6 & {100.0} & \textbf{77.9} \\
\bottomrule
\end{tabular}
}
\label{tab:stressid_subject_full_metrics}
\end{table*}

\begin{table*}[t]
\centering
\setlength{\tabcolsep}{3pt}
\renewcommand{\arraystretch}{1.15}

\caption{\textbf{Per-subject performance across 10 BAH target subjects.}
Across three matrices \war, \uar, and \fonescore metrics across 10 subjects.}

\resizebox{\linewidth}{!}{%
\begin{tabular}{lccccccccccc}
\toprule
\textbf{Method} & Sub-1 & Sub-2 & Sub-3 & Sub-4 & Sub-5 & Sub-6 & Sub-7 & Sub-8 & Sub-9 & Sub-10 & {Avg} \\
\midrule
\multicolumn{12}{l}{\textbf{\war (\%)}} \\
\midrule
TPT & 62.0 & 53.7 & 73.9 & 80.0 & 81.2 & 71.0 & 58.0 & 55.0 & 69.0 & 53.0 & 65.6 \\
TDA & 58.0 & 53.7 & 75.0 & 80.3 & 79.0 & 71.6 & 58.0 & 55.9 & 69.0 & 51.7 & 65.2 \\
DPE & 58.6 & 59.6 & 79.6 & 80.5 & 83.4 & 67.0 & 59.0 & 54.0 & 71.9 & 54.0 & 66.7 \\
PromptAlign & 59.1 & 59.1 & 72.0 & 84.5 & 84.1 & 72.4 & 58.5 & 55.9 & 71.5 & 54.9 & 67.1 \\
ReTA & 61.4 & 60.0 & 72.0 & 84.0 & 84.1 & 72.4 & 58.5 & 55.9 & 72.0 & 56.0 & 67.6 \\
T3AL & 61.0 & 58.9 & 75.0 & 85.6 & 83.6 & 74.7 & 60.0 & 56.0 & 71.0 & 53.6 & 67.9 \\
\rowcolor{gray!15}\textbf{\ourstt} & 54.7 & 61.3 & 87.2 & 88.4 & 85.9 & 76.4 & 61.6 & 56.7 & 72.0 & 54.1 & \textbf{69.8} \\
\midrule
\multicolumn{12}{l}{\textbf{\uar (\%)}} \\
\midrule
TPT & 51.0 & 50.0 & 50.0 & 50.0 & 51.0 & 53.1 & 50.0 & 49.0 & 50.0 & 50.0 & 50.4 \\
TDA & 50.0 & 50.0 & 50.0 & 51.0 & 50.0 & 50.0 & 50.0 & 50.0 & 50.0 & 50.0 & 50.1 \\
DPE & 50.0 & 50.0 & 50.0 & 50.0 & 50.0 & 50.0 & 50.0 & 50.0 & 50.0 & 50.0 & 50.0 \\
PromptAlign & 50.0 & 50.0 & 50.2 & 51.8 & 50.0 & 50.0 & 50.0 & 50.0 & 50.0 & 50.0 & 50.2 \\
ReTA & 50.0 & 50.0 & 50.2 & 51.8 & 50.0 & 50.0 & 50.0 & 50.0 & 50.0 & 50.0 & 50.2 \\
T3AL & 51.4 & 50.0 & 50.0 & 50.0 & 51.4 & 54.0 & 54.3 & 50.0 & 50.0 & 50.0 & \textbf{51.1} \\
\rowcolor{gray!15}\textbf{\ourstt} & 52.0 & 50.0 & 50.0 & 50.0 & 50.0 & 50.0 & 50.0 & 50.0 & 50.0 & 50.0 & {50.2} \\
\midrule
\multicolumn{12}{l}{\textbf{\fonescore (\%)}} \\
\midrule
TPT & 40.6 & 35.7 & 43.7 & 45.9 & 45.6 & 43.1 & 35.0 & 34.7 & 40.7 & 32.9 & 39.7 \\
TDA & 40.0 & 39.7 & 44.4 & 46.0 & 42.0 & 42.6 & 35.0 & 35.5 & 40.7 & 33.2 & 39.9 \\
DPE & 34.8 & 37.1 & 44.2 & 46.0 & 45.3 & 43.0 & 37.2 & 32.5 & 41.0 & 33.2 & 39.4 \\
PromptAlign & 25.0 & 37.0 & 46.9 & 51.0 & 46.1 & 43.5 & 41.0 & 32.5 & 41.0 & 33.2 & 39.7 \\
ReTA & 25.0 & 36.9 & 41.9 & 51.0 & 46.1 & 51.2 & 40.0 & 32.5 & 41.0 & 33.2 & 39.8 \\
T3AL & 41.6 & 37.0 & 44.7 & 46.0 & 45.2 & 43.0 & 38.0 & 36.2 & 41.0 & 35.0 & 40.7 \\
\rowcolor{gray!15}\textbf{\ourstt} & 38.8 & 38.0 & 46.6 & 46.9 & 46.2 & 43.3 & 38.1 & 36.2 & 41.9 & 35.1 & \textbf{41.1} \\
\bottomrule

\end{tabular}
}
\label{tab:bah_subject_full_metrics_corrected}
\end{table*}

\begin{table}[t]
\centering
\small
\setlength{\tabcolsep}{5pt}
\renewcommand{\arraystretch}{1.1}
\caption{We evaluate different prompt templates for representing emotion classes in the CLIP text encoder. The best-performing template, \emph{``a person with an expression of [CLS]''}, is used for all methods that rely on class prompts (CPs).}
\begin{tabular}{lcc}
\toprule
\textbf{Prompt Template} & \textbf{\war} & \textbf{\fonescore} \\
\midrule
a photo of a [CLS]                                      & 68.7 & 63.3 \\
a\_photo\_of\_the\_[CLS]\_face                          & 66.5 & 59.3 \\
a\_photo\_of\_one\_[CLS]\_face                          & 64.2 & 55.8 \\
a\_close-up\_photo\_of\_the\_[CLS]\_face                & 68.2 & 62.6 \\
a\_low\_resolution\_photo\_of\_a\_[CLS]\_face           & 68.7 & 64.2 \\
a\_good\_photo\_of\_a\_[CLS]\_face                      & 69.2 & 65.4 \\
a\_photo\_of\_my\_[CLS]\_face                           & 66.2 & 60.3 \\
a\_cropped\_photo\_of\_the\_[CLS]\_face                 & 64.5 & 57.5 \\
a\_photo\_of\_a\_person\_with\_[CLS]\_face              & 67.2 & 61.4 \\
\rowcolor{gray!15}
\textbf{a person with an expression of [CLS]}           & \textbf{69.7} & \textbf{66.6} \\
a\_portrait\_of\_a\_person\_in\_[CLS]                   & 58.2 & 46.7 \\
a\_face\_showing\_signs\_of\_[CLS]                      & 68.7 & 64.1 \\
a close up portrait of a [CLS] expression               & 69.2 & 63.0 \\
a realistic photo of a person experiencing [CLS]        & 53.5 & 40.0 \\
a cropped image of a person in [CLS]                    & 69.0 & 64.5 \\
a photo of a [CLS] person                               & 69.5 & 64.3 \\
a person with a facial expression of [CLS]              & 69.5 & 64.2 \\
a high quality photo of a [CLS] expression              & 69.5 & 64.3 \\
a photo of a face showing [CLS]                         & 69.5 & 64.3 \\
a photo of a face in [CLS]                              & 69.0 & 61.8 \\
\bottomrule
\end{tabular}
\label{tab:class_prompt_ablation_full}
\end{table}

\begin{table*}[t]
\centering
\small
\setlength{\tabcolsep}{5pt}
\renewcommand{\arraystretch}{1.05}
\caption{
\textbf{Full list of Class Prompts (CPs) used in the ablation study.}
Each emotion category replaces \textbf{[CLS]} with the target label (e.g., \textit{pain}, \textit{stress}, or \textit{ambivalence}).
We employ 46 templates. 23 emotion-specific and 23 neutral, to capture a broad linguistic and visual context.
The prompts follow CLIP-style phrasing (\textit{``a photo of a [CLS] face''}, \textit{``a person with an expression of [CLS]''}), ensuring alignment with large-scale vision-language pretraining.
}
\begin{tabular}{p{0.47\linewidth} p{0.47\linewidth}}
\toprule
\textbf{Emotion Prompts ([CLS])} & \textbf{Neutral Prompts} \\
\midrule
a photo of a \textbf{[CLS]} face & a photo of a neutral face \\
a photo of the \textbf{[CLS]} face & a photo of the neutral face \\
a photo of one \textbf{[CLS]} face & a photo of one neutral face \\
a close-up photo of the \textbf{[CLS]} face & a close-up photo of the neutral face \\
a low resolution photo of a \textbf{[CLS]} face & a low resolution photo of a neutral face \\
a good photo of a \textbf{[CLS]} face & a good photo of a neutral face \\
a photo of my \textbf{[CLS]} face & a photo of my neutral face \\
a cropped photo of the \textbf{[CLS]} face & a cropped photo of the neutral face \\
a photo of a person with a \textbf{[CLS]} face & a photo of a person with a neutral face \\
a photo of a face with an expression of \textbf{[CLS]} & a photo of a face with an expression of neutral \\
a person with an expression of \textbf{[CLS]} & a person with an expression of neutral \\
a portrait of a person showing \textbf{[CLS]} & a portrait of a person with a neutral expression \\
a face showing signs of \textbf{[CLS]} & a face showing no expression \\
a close-up portrait of a \textbf{[CLS]} expression & a close-up portrait of a neutral expression \\
a realistic photo of a person experiencing \textbf{[CLS]} & a realistic photo of a person with a neutral face \\
a photo of a \textbf{[CLS]} expression on a face & a photo of a blank expression on a face \\
a cropped image of a person in \textbf{[CLS]} & a cropped image of a person with a neutral face \\
a facial expression indicating \textbf{[CLS]} & a facial expression indicating no emotion \\
a photo of a person feeling \textbf{[CLS]} & a photo of a calm person \\
a person with a facial expression of \textbf{[CLS]} & a person with a relaxed facial expression \\
a high quality photo of a \textbf{[CLS]} expression & a high quality photo of a neutral expression \\
a photo of a face showing \textbf{[CLS]} & a photo of a face in rest state \\
a photo of a face in \textbf{[CLS]} & a photo of a face with no expression \\
\bottomrule
\end{tabular}
\label{tab:class_prompts_all}
\end{table*}

\subsection{Action Units Used in \textbf{\ourstt}}
\label{subsec:supp_aus}

Action Units (AUs) originate from the Facial Action Coding System (FACS), which decomposes facial behaviour into atomic muscle activations~\cite{ekman1978facial}. Each AU corresponds to specific facial muscle movements (e.g., \textit{inner brow raise}, \textit{lip corner pull}, \textit{eye tightening}), providing an anatomically grounded and interpretable representation of facial expressions. Since emotions arise from combinations of these localized activations, AUs offer a rich mid-level representation capable of capturing subtle expressive cues often missed by categorical emotion labels. In \ourstt, we adopt a set of 46 AUs widely used in AU detection and expression analysis~\cite{martinez2017automatic,belharbi2024guided,zhang2024multimodal}. These AUs cover eyebrow, eyelid, cheek, nose, lip, and jaw movements commonly observed in pain, stress, and subtle expression datasets. Each AU is converted into a textual prompt, enabling CLIP to align facial dynamics with semantically meaningful and anatomically interpretable units. This AU space forms the basis for both AU-guided pretraining (\ours) and the test-time personalization module (\ourstt). Tab.~\ref{tab:au_prompts_full} lists all 46 AUs used in our experiments.

\subsection{Additional Personalization Results}
\label{subsec:supp_personalization}
We report personalization results across 10 target subjects for the {BioVid}, {StressID}, and {BAH} datasets. For each subject, we provide \war, \uar, and \fonescore to enable fine-grained analysis of subject-specific behavior under subtle and heterogeneous facial expressions.
\textbf{BioVid} contains subtle pain expressions with pronounced inter-subject variation in AU activation patterns. Subject-wise results in Tab.~\ref{tab:biovid_subject_full_metrics} show that existing CLIP-based TTA baselines fluctuate substantially, particularly for subjects with weak or inconsistent pain cues. In contrast, \ourstt delivers the most stable and highest average performance across all three metrics. \textbf{StressID} consists of spontaneous and fine-grained stress behaviors, which vary widely across individuals and are often influenced by cognitive and physiological state changes. The baselines exhibit significant instability on subjects whose stress cues are subtle or temporally inconsistent (Tab.~\ref{tab:stressid_subject_full_metrics}). \ourstt consistently achieves superior performance across all subjects, highlighting its ability to personalize AU semantics and capture weak, subject-dependent stress signals. \textbf{BAH} includes extremely subtle and ambivalent facial behaviors, making it the most challenging personalization setting. Although \ourstt attains the best average performance across \war, \uar, and \fonescore as shown in Tab.~\ref{tab:bah_subject_full_metrics_corrected}, the dataset reveals an inherent limitation of AU-based personalization. Ambivalent expressions such as hesitation or uncertainty often manifest through overlapping or contradictory AU combinations that differ substantially between individuals. This variability makes it difficult to infer a stable AU activation pattern, and when only short video segments are available, the model may over-adapt to transient cues.

\begin{figure*}[t!]
\centering
\includegraphics[width=1.0\linewidth]{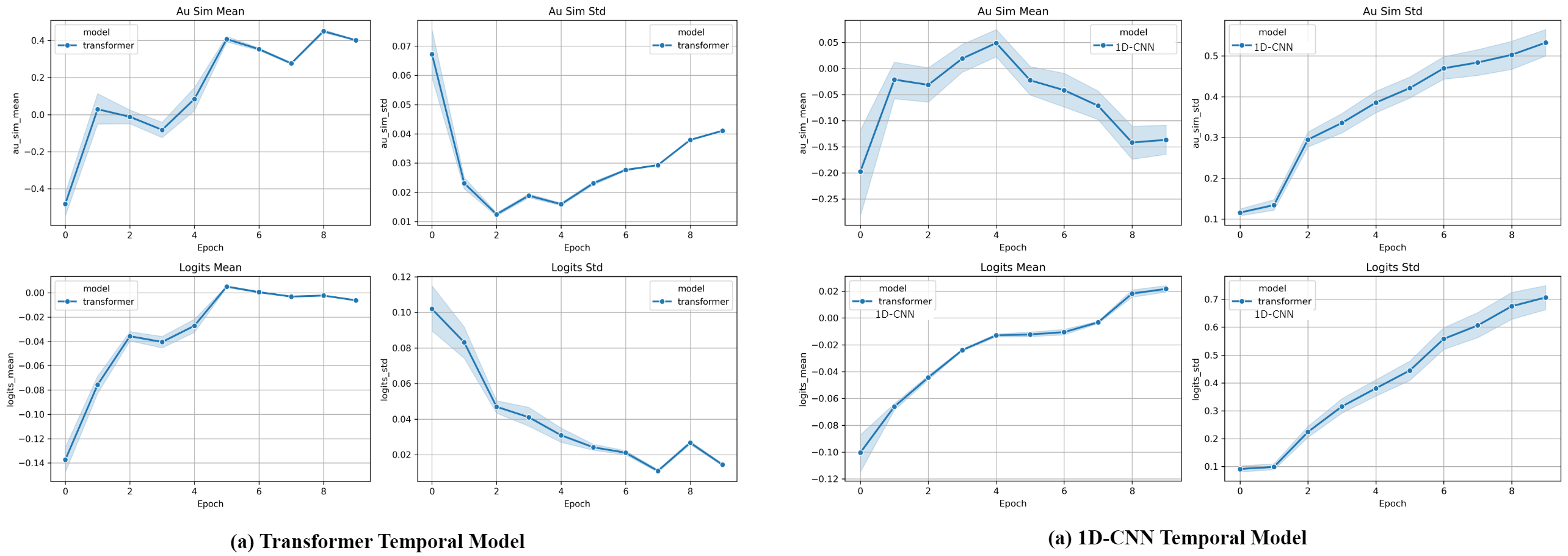}
\caption{\textbf{AU–video similarity statistics.} 
Comparison of the mean and standard deviation of AU–video similarity logits over $N$ epochs for 
(a) Transformer temporal module and 
(b) 1D-CNN temporal module.}
\label{fig:temporal_dc}      
\vspace{-10pt}
\end{figure*}

\begin{figure}[t!]
\centering
\includegraphics[width=0.8\linewidth]{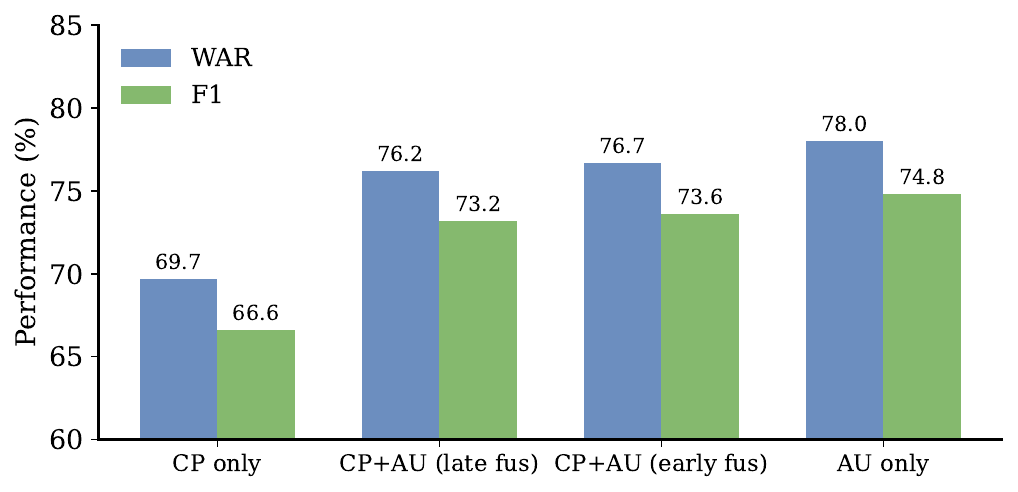}
\caption{
Comparison of class prompt (CP), action unit (AU), and combined CP+AU performance on the BioVid dataset. 
}
\label{fig:class_prompt_ablation}      
\vspace{-10pt}
\end{figure}

\section{Ablation}
\label{sec:supp_ablation}

\subsection{Temporal Module Design Choice}
\label{subsec:supp_temporal}

To study temporal modeling for fine-grained FER, we first evaluate a lightweight transformer module, motivated by its success in basic emotion recognition. However, as shown in Fig.~\ref{fig:temporal_dc}(a), AU–video similarity and logit statistics across epochs reveal unstable dynamics when applied to subtle expressions. The AU similarity mean fluctuates significantly, while the standard deviation collapses and later rises again, indicating overfitting to short-range temporal noise and inconsistent AU-driven cues. Logit statistics also show reduced variance and weak activation magnitudes, suggesting limited ability to amplify discriminative temporal evidence. To address this issue, we introduce a 1D-CNN temporal encoder with Gated Linear Units (GLU) and temporal pooling (Fig.~\ref{fig:temporal_dc}(b)). The convolutional layers capture local temporal dynamics, while GLU selectively gates informative channels. This design produces stable AU similarity trends, where the mean steadily increases and the standard deviation grows smoothly, indicating stronger AU alignment and improved discriminability. Logits also become more separated and confident across epochs. Overall, the 1D-CNN+GLU module better captures the micro-temporal variations of subtle expressions, providing a reliable foundation for personalized AU–video alignment in \ours.

\begin{figure*}[t!]
\centering
\includegraphics[width=1.0\linewidth]{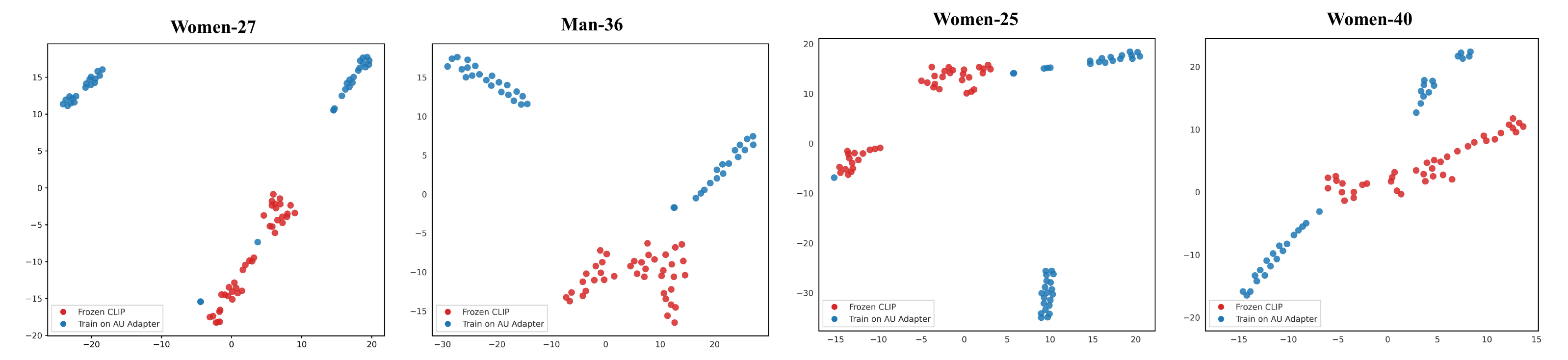}
\caption{
t-SNE visualization of video–AU similarity space between \textcolor{red}{frozen clip} and \textcolor{blue}{AU adapter}.}
\label{fig:au_tsne}      
\vspace{-10pt}
\end{figure*}

\subsection{Class-Prompt Template Selection}
\label{subsec:supp_class_template}
To evaluate the influence of textual formulations in CLIP-based FER, we conduct an ablation over a diverse set of natural-language class prompts that are commonly used in FER and CLIP prompt-engineering studies~\cite{zhang2023learning, foteinopoulou2024emoclip, zhao2025enhancing}.
Since subtle-expression personalization relies on fine-grained alignment between text prompts and subject-specific appearance cues, it is important to identify whether certain templates provide more stable or discriminative supervision.
We evaluate different prompt templates as shown in Tab.~\ref{tab:class_prompt_ablation_full}, ranging from simple category labels (e.g., “a photo of a [CLS]”) to more descriptive expressions (e.g., “a face showing signs of [CLS]”). Overall, most templates yield similar performance. However, one formulation, “a person with an expression of [CLS]” consistently achieves the strongest results across \war, \uar, and \fonescore, especially in subtle-expression settings. Based on this observation, we adopt \emph{“a person with an expression of [CLS]”} as the default class prompt for all methods that rely on class prompts (CPs) in our experiments.

\subsection{Replacing AU Prompts with Class Prompts}
\label{subsec:supp_class_replace}
In the main paper, Sec. 4.4 (Are Action Units Really Helping?), we perform an ablation study by using an ensemble of CPs to train our AU adapter and emotion classifier. To ensure a fair comparison with the AU-based representations, we constructed 46 textual prompts, 23 for neutral and 23 for the corresponding target emotion (e.g., pain, stress, or ambivalence) as shown in Tab.~\ref{tab:class_prompts_all}. 
Each prompt follows natural language structures inspired by CLIP-style pretraining data (e.g., \textit{“a photo of a person with an expression of pain”}, \textit{“a close-up portrait of a neutral face”}). 
This design allows us to evaluate whether a diverse set of class-level textual descriptions alone can substitute for the physiological cues encoded by AUs. 

\subsection{Effect of adding Class-prompts with AU-prompts}
\label{subsec:supp_cp_au}
We evaluate the integration of class prompts (CP) with AU-based representations (CP+AU). The class prompt follows the template \emph{``a person with an expression of [CLS]''}, where \texttt{[CLS]} denotes \texttt{Neutral} or \texttt{Pain}. For fair comparison, CP-only and CP+AU use the same CLIP backbone pre-trained on the source domain (as in EmoCLIP), identical to our AU-only setup. For CP+AU (late fusion), AU logits are combined with CP logits computed via cosine similarity between class prompts and visual embeddings. In CP+AU (early fusion), AU–visual and CP–visual similarities are concatenated before the classifier. As shown in Fig.~\ref{fig:class_prompt_ablation}, early fusion slightly improves over late fusion, but both remain inferior to AU-only, indicating that AU prompting provides the most discriminative representation by focusing on localized facial activations.

\subsection{Impact of AU Adapter}
\label{subsec:supp_adapter}
Fig.~\ref{fig:au_tsne} visualizes the t-SNE projections of video–AU similarity vectors for four subjects (\texttt{Women-27}, \texttt{Man-36}, \texttt{Women-25}, and \texttt{Women-40}) before and after adding the AU adapter 
. In the frozen CLIP model (\textcolor{red}{red}), the similarity patterns are highly overlapping and weakly structured, indicating limited discrimination among AU semantics. After training with the AU adapter (\textcolor{blue}{blue}), the representations become distinctly separated and form coherent clusters, showing that the AU adapter strengthens the alignment between visual features and AU-specific activations. This consistent improvement across subjects demonstrates that adding the AU adapter produces more meaningful and discriminative AU–video similarity representations.

\subsection{Extended Key-Frame Selection Analysis}
\label{subsec:supp_keyframe}
\begin{figure}[t!]
\centering
\includegraphics[width=0.8\linewidth]{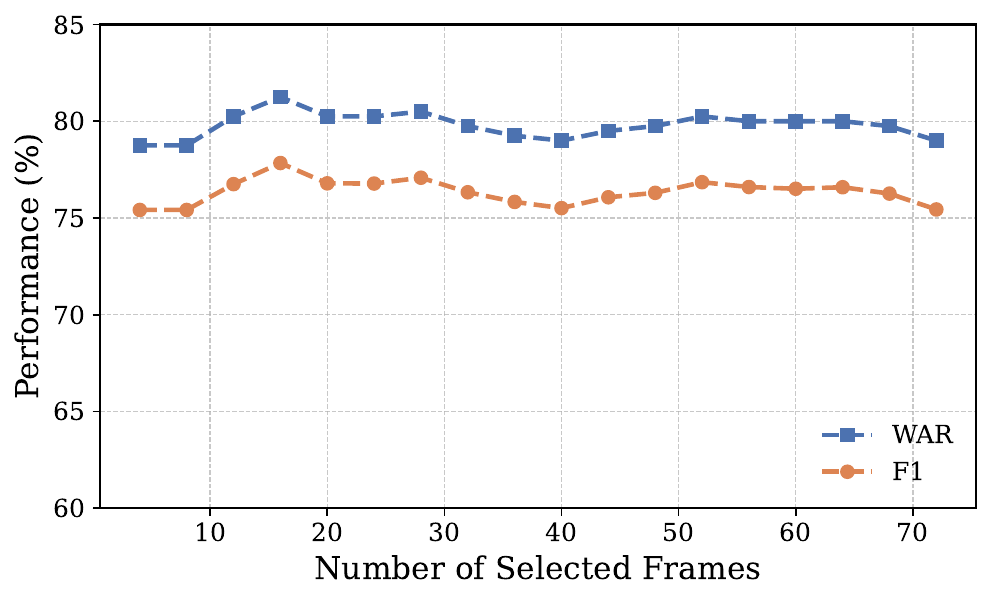}
\caption{Ablation of temporal window length. A 16-frame window provides the best trade-off between accuracy and efficiency.}
\label{fig:key_frame_sel}      
\vspace{-10pt}
\end{figure}

In addition to the ablation reported in the main paper Sec. 4.4 (Impact of Temporal Window Selection.), we provide an extended evaluation of the key-frame selection as shown in Fig~\ref{fig:key_frame_sel} on the BioVid dataset. This experiment examines a wider range of temporal sampling intervals, selecting consecutive low-entropy frames from \{4, 8, 12, \dots, 72\} and reporting both \war and \fonescore for each configuration. The expanded results confirm the trends observed in the main paper. Selecting a smaller number of informative frames consistently preserves or improves performance, with 16 frames offering the best balance between accuracy and efficiency. More importantly, the method remains largely insensitive to the precise number of selected frames, as \ourstt reliably identifies temporally coherent, low-entropy segments that capture the most expressive and discriminative facial cues.

\subsection{Additional Analysis of \ourstt}
\label{subsec:supp_additional}

\noindent\textbf{(a) Robustness of \ourstt.} To assess robustness across random initialization, we repeat the \stressid experiments over three random seeds. Figure~\ref{fig:robustness_window}~(a) reports the mean and standard deviation for T3AL and \ourstt. The consistent improvements in both \war and \fonescore confirm that the gains of \ourstt are stable and not driven by a single run.

\noindent\textbf{(b) Window selection fairness and full-video comparison.} Window selection is part of \ourstt, but we apply the same protocol to the TTA baseline, using T3AL as the video-level variant. We also add a full-frame ablation using all frames. As shown in the Figure~\ref{fig:robustness_window}~(b), \ourstt outperforms T3AL in both settings, showing that the gain is not solely due to window selection. Only \war is shown due to space limits.

\begin{figure}[H]
\centering
\begin{minipage}[c]{0.35\linewidth}
    \centering
    \scriptsize
    \setlength{\tabcolsep}{3.5pt}
    \begin{tabular}{lcc}
        \toprule
        Method & \war & \fonescore \\
        \midrule
        T3AL
        & $75.9 \pm 0.13$
        & $58.5 \pm 0.81$ \\
        \ourstt
        & $\mathbf{80.3 \pm 0.75}$
        & $\mathbf{77.1 \pm 0.75}$ \\
        \bottomrule
    \end{tabular}
\end{minipage}
\hspace{0.05\linewidth}
\begin{minipage}[c]{0.40\linewidth}
    \centering
    \setlength{\tabcolsep}{8pt}
    \begin{tabular}{lcc}
        \toprule
        Method & Win. Sel. & All \\
        \midrule
        T3AL & 77.0 & 75.2 \\
        \ourstt & \textbf{81.5} & \textbf{79.5} \\
        \bottomrule
    \end{tabular}
\end{minipage}


\vspace{-4pt}
\caption{
Additional robustness analysis: mean and standard deviation on \stressid over three random seeds, and candidate temporal windows with
the selected minimum-entropy window.
}
\label{fig:robustness_window}
\vspace{-8pt}
\end{figure}

\begin{figure}[H]
\centering
\begin{minipage}[c]{0.80\linewidth}
    \centering
    \includegraphics[width=\linewidth]{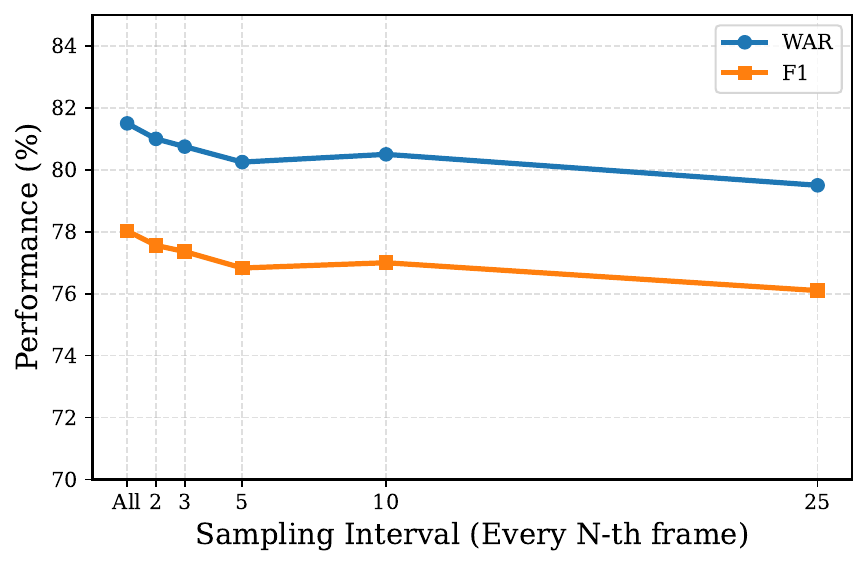}
\end{minipage}
\hfill
\caption{BioVid analysis on different frame-sampling rate.}
\label{fig:prompt_sampling}
\end{figure}

\noindent\textbf{(c) Different sampling rate and video encoder} 
We added a controlled BioVid sampling study as shown in  Figure~\ref{fig:prompt_sampling}. All frames perform best, while sparse sampling provides an efficiency trade-off. We keep frozen CLIP for AU-text/visual alignment, as video encoders require extra alignment or fine-tuning and could confound gains.

\noindent\textbf{(d) \ourstt on common FER datasets.}
Although our focus is fine-grained video ER for health-related applications, we also evaluate on \text{FERV39K}, as suggested. As shown in the As shown in Table~\ref{tab:supp_combined_analysis}~(a), \ours outperforms the CLIP-FER baseline, while \ourstt surpasses the TTA baseline, demonstrating generalization beyond subtle ER datasets.

\noindent\textbf{(e) AU prompts versus sentence-level prompts} As shown in Table~\ref{tab:supp_combined_analysis}~(b), we change the 46 phrase-level to sentence-level descriptions as in SEV-Net. Both \ours and \ourstt use the matched prompt style to avoid train-test mismatch. Sentence-level prompts underperform, suggesting concise AU phrases provide cleaner localized anchors for AU-video alignment.

\begin{table}[H]
\centering
\scriptsize
\begin{tabular}{@{}c@{\hspace{0.06\linewidth}}c@{}}

\begin{tabular}{llcc}
\toprule
Group & Method & \war & \fonescore \\
\midrule
\multirow{2}{*}{{CLIP-FER}}
& EmoCLIP & 41.5 & 36.6 \\
& \ours & 47.3 & 41.4 \\
\midrule
\multirow{2}{*}{{TTA}}
& T3AL & 43.8 & 38.1 \\
& \ourstt & \textbf{51.1} & \textbf{44.7} \\
\bottomrule
\end{tabular}
&
\begin{tabular}{llcc}
\toprule
Prompt & Method & \war & \fonescore\\
\midrule
\multirow{2}{*}{{Phrase-level}}
& \ours & \textbf{78.0} & \textbf{74.8} \\
& \ourstt & \textbf{81.5} & \textbf{78.0} \\
\midrule
\multirow{2}{*}{{Sentence-level}}
& \ours & 73.2 & 69.5 \\
& \ourstt & 78.5 & 75.2 \\
\bottomrule
\end{tabular}
\end{tabular}
\vspace{5pt}
\caption{(a) comparison of CLIP-FER and TTA methods on FERV39K, (b) phrase-level versus sentence-level AU prompts on BioVid.}
\label{tab:supp_combined_analysis}
\end{table}

\clearpage
\putbib[main]
\end{bibunit}

\end{document}